\documentclass[12pt,letterpaper]{article}
\usepackage[a4paper, total={7in, 10in}]{geometry}

\usepackage{helvet}
\usepackage{authblk}
\usepackage{hyperref}

\usepackage{amsmath}
\usepackage{booktabs}
\usepackage{graphicx}
\usepackage{float} 
\usepackage{tabularx}
\usepackage{amssymb}

\makeatletter
\renewcommand{\maketitle}{\bgroup\setlength{\parindent}{0pt}
\begin{flushleft}
  \textbf{\@title}
  
  \@author
\end{flushleft}\egroup}
\makeatother

\title{Bridging the gap between Performance and Interpretability: An Explainable Disentangled Multimodal Framework for Cancer Survival Prediction}
\date{}

\author[1,*]{Aniek Eijpe}
\author[1,2,3]{Soufyan Lakbir}
\author[4]{Melis Erdal Cesur}
\author[4]{Sara P. Oliveira}
\author[5]{Angelos Chatzimparmpas}
\author[1]{Sanne Abeln}
\author[1,6,*]{Wilson Silva}

\affil[1]{AI Technology for Life, Department of Information and Computing Sciences, Department of Biology, Utrecht University, Utrecht, The Netherlands}
\affil[2]{Department of Metabolic Diseases, Wilhelmina Children’s Hospital, University Medical Center Utrecht, Utrecht, The Netherlands} \affil[3]{Regenerative Medicine Center Utrecht, Utrecht, The Netherlands}
\affil[4]{Computational Pathology, Department of Pathology, The Netherlands Cancer Institute, Amsterdam, The Netherlands}
\affil[5]{Visualization and Graphics, Department of Information and Computing Sciences, Utrecht University, Utrecht, The Netherlands}
\affil[6]{Lead contact}

\affil[*]{a.eijpe@uu.nl, w.j.dossantossilva@uu.nl}

\usepackage[super,comma,sort&compress]{natbib}

\begin{document}

\maketitle

\section*{Summary}
While multimodal survival prediction models are increasingly more accurate, their complexity often reduces interpretability, limiting insight into how different data sources influence predictions. To address this, we introduce DIMAFx, an explainable multimodal framework for cancer survival prediction that produces disentangled, interpretable modality-specific and modality-shared representations from histopathology whole-slide images and transcriptomics data. Across multiple cancer cohorts, DIMAFx achieves state-of-the-art performance and improved representation disentanglement. Leveraging its interpretable design and SHapley Additive exPlanations, DIMAFx systematically reveals key multimodal interactions and the biological information encoded in the disentangled representations. In breast cancer survival prediction, the most predictive features contain modality-shared information, including one capturing solid tumor morphology contextualized primarily by late estrogen response, where higher-grade morphology aligned with pathway upregulation and increased risk, consistent with known breast cancer biology. Key modality-specific features capture microenvironmental signals from interacting adipose and stromal morphologies. These results show that multimodal models can overcome the traditional trade-off between performance and explainability, supporting their application in precision medicine.

\section*{Keywords}
Cancer survival prediction, Disentangled representation learning, Multimodal deep learning, Interpretability and explainability in AI, Histopathology whole-slide images, Transcriptomics data

\section*{Introduction}

Cancer presents a significant global health challenge, with a critical need for improved patient risk stratification~\cite{force2025global}.
Multimodal models have rapidly emerged to address this challenge~\cite{li2022hfbsurv,mcat,mmp,survpath,motcat,cmta,yan2025pathway,qu2025multimodal,pibd,dimaf}, driven by advances in deep learning and the increasing availability of high-dimensional data, such as transcriptomics data or histopathology whole-slide images (WSIs).
Transcriptomics data can capture pathway signatures and molecular subtypes, such as the breast cancer subtypes (\textit{e.g.}, Luminal A, Luminal B, HER2-enriched, and Basal-like)~\cite{sorlie2001gene,cancer2012comprehensive}, that have distinct biology, prognosis, and therapy response profiles. In contrast, WSIs contain tissue architecture, tumor morphology, and micro-environmental patterns that are not directly observable from molecular data alone~\cite{chen2022pan,mani2022cancer}.
By integrating these heterogeneous data sources, multimodal systems have consistently demonstrated improved performance over unimodal models~\cite{li2022hfbsurv,survpath,mcat,motcat,cmta,qu2025multimodal,mmp,pibd,yan2025pathway}.

These performance gains generally rely on sophisticated fusion strategies that capture local cross-modal interactions~\cite{li2022hfbsurv,motcat,cmta,mmp}, blending the information from both modalities into a highly entangled latent space~\cite{pibd,cui2025premise}.
While effective, this also increases model complexity and obscures the nature of the information encoded in the learned multimodal representations, illustrating a critical interpretability gap~\cite{lipton2018mythos,zhang2021survey,murdoch2019definitions}. 
This lack of interpretability can undermine trust, hinder adoption, or lead to serious consequences in a clinical setting~\cite{kundu2021ai,rudin2019stop}, while also limiting scientific discovery by obscuring the biological patterns learned from the data~\cite{zhang2021survey}.
Moreover, regulatory frameworks, including the European Union’s Ethics Guidelines for Trustworthy AI and the AI Act (Regulation (EU) 2024/1689), stress transparency for high-risk medical AI systems~\cite{EU_guide,EU_AI_Act_Article13}. 
Developing inherently interpretable models satisfies critical clinical and scientific requirements while supporting the transparency emphasized by these regulations.
While few multimodal survival models incorporate interpretability or explainability, they typically focus only on the unimodal features or selected cross-modal interactions of a few cases~\cite{mmp,yan2025pathway,chen2022pan,survpath}.
This leaves the multimodal representations largely opaque, obscuring which information is shared or complementary across modalities and how interactions between or within modalities drive predictions.

Disentangled Representation Learning (DRL)~\cite{bengio2013representation,higgins2018towards} offers a promising solution to improve interpretability by explicitly separating the modality-specific and modality-shared information in disentangled representations. This yields more structured representations while also reducing the risk of shared signals suppressing complementary, modality-specific information~\cite{pibd,dimaf}. In the context of multimodal survival analysis, DRL-based approaches have promoted disentanglement through architectural design~\cite{fu2025hsfsurv} and unsupervised objectives such as orthogonality constraints~\cite{wu2023camr}, contrastive loss functions~\cite{robinet2024drim}, adversarial training~\cite{wu2023camr,long2024mugi}, or mutual information-based approximations~\cite{pibd}. 
Recently, we introduced Disentangled and Interpretable Multimodal Attention Fusion (DIMAF)~\cite{dimaf}, a framework that integrates WSIs and transcriptomics data through disentangled attention-based fusion to create separate modality-specific and modality-shared representations. DIMAF further promotes linear and nonlinear disentanglement through a loss based on Distance Correlation (DC)~\cite{empericaldcor,liu2020metrics}, showing that it is possible to obtain state-of-the-art performance with disentangled representations. However, while these methods yield more structured representations, they provide no further insight into which biological signals the modality-specific and modality-shared features encode and how they influence predictions.

To address this, we propose DIMAFx, an \textit{explainable} framework that integrates WSIs and transcriptomics data for cancer survival prediction. DIMAFx adopts the disentanglement strategy of DIMAF, separating the intra- and inter-modal interactions and leveraging the DC-based loss to produce disentangled modality-specific and modality-shared representations.  
Addressing the limitations of our previous work, DIMAFx embeds these disentangled representations into a unified explainable framework, enabling an in-depth interpretability analysis to examine the biological information encoded in the unimodal features, their interactions, and in the resulting modality-shared and modality-specific representations. To support this, WSI representations are enhanced to better capture the frequency of morphological patterns. Moreover, by using SHapley Additive exPlanations (SHAP)~\cite{NIPS2017_7062}, DIMAFx can quantify the contribution of the unimodal and multimodal features to survival prediction, identifying key drivers of risk.
DIMAFx also improves representation disentanglement over DIMAF by introducing a learnable aggregation layer that adaptively combines the most informative modality-specific and modality-shared features.

We demonstrate that DIMAFx achieves state-of-the-art survival prediction, risk stratification, and improved disentanglement on TCGA data collections for breast, bladder, lung, and kidney carcinomas. Moreover, we conducted an extensive case study on the breast cancer cohort, showing that DIMAFx captures biologically meaningful patterns consistent with known biological mechanisms in breast cancer. The most predictive features were modality-shared, including one representing KRAS signaling contextualized by tumor morphological features and another capturing solid tumor morphology contextualized primarily by the late estrogen response pathway. 
Among the modality-specific features, interacting adipose and stromal morphologies were most influential, highlighting predictive signals from the tumor microenvironment. 
These results show that DIMAFx provides strong interpretability without compromising performance, effectively bridging the traditional explainability gap.

\section*{Results}

\subsection*{Learning structured and interpretable multimodal features}
\begin{figure}[H]
    \centering
    \includegraphics[width=\linewidth, trim=0 45 0 30]{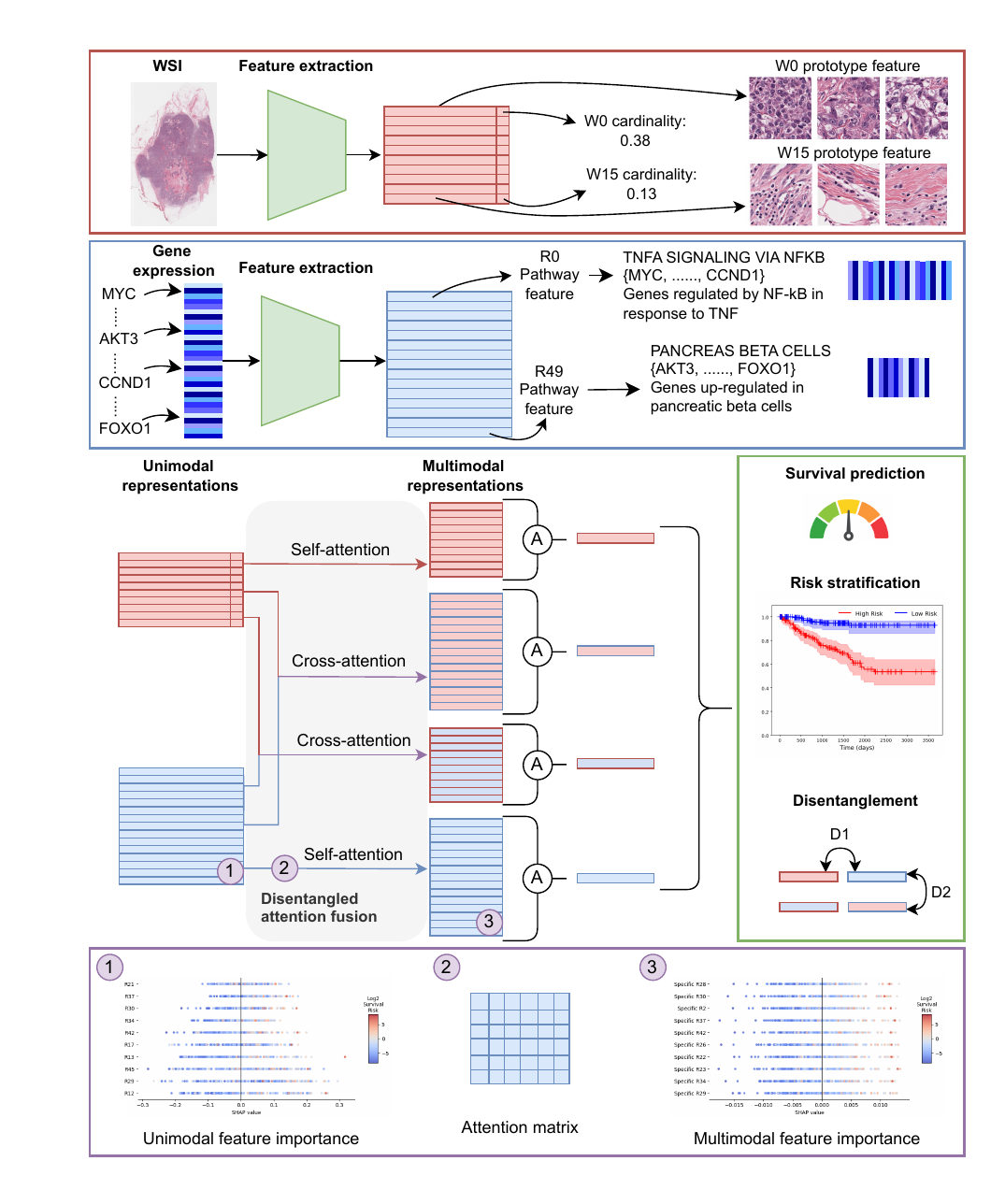}
    \caption{\textbf{Overview of DIMAFx}. Our proposed framework encodes each modality with interpretable features, then applies disentangled attention fusion and aggregation to produce four disentangled representations. The representations are used for the downstream tasks, including survival prediction. Both unimodal and multimodal feature importance is assessed using SHAP values, and the attention matrices from the self- and cross-attention layers provide insights into the intra- and inter-modal interactions.}
    \label{fig:DIMAFx}
\end{figure}

DIMAFx (Figure~\ref{fig:DIMAFx}) represents each modality using inherently interpretable features: morphological prototype features for WSIs~\cite{panther,mmp} and pathway-level features to represent the gene expression data (\textit{i.e.}, transcriptomics data)~\cite{survpath,mmp,pibd}. The WSI prototype features are interpretable by visualizing the patches most similar to each prototype, revealing the associated morphological patterns, and by examining their cardinality, reflecting their frequency within a slide (see red box). In contrast, the transcriptomic pathway features are interpretable by design because each corresponds to a predefined gene set with known biological function (as indicated in the blue box).

Using these unimodal features, DIMAFx models the intra- and inter-modal interactions separately using disentangled attention-based fusion~\cite{dimaf}, resulting in four distinct multimodal representations, as shown in Figure~\ref{fig:DIMAFx}(middle): a WSI-specific representation (top), a transcriptomics-specific representation (bottom), and two shared representations (middle two). The shared representations encode the shared information across both modalities, one where the WSI features are contextualized by the transcriptomics features (bottom) and one where the transcriptomics features are contextualized by the WSI features (top). Although the modality-specific representations are not strictly multimodal, we refer to all four disentangled representations collectively as the multimodal representations for simplicity.
DIMAFx further promotes disentanglement by penalizing dependence between the modality-specific representations (D1) and between the modality-specific and the modality-shared representations (D2), thereby effectively capturing complementary modality-specific signals separately from the modality-shared information.

A key novelty of DIMAFx is its explainability: each unimodal and multimodal latent feature has a clearly defined biological semantic meaning, and SHAP analysis enables us to quantify their contributions to patient risk (see Figure~\ref{fig:DIMAFx} purple box (1) and (3), respectively). Additionally, the learned attention matrices (purple box (2)) reveal the patterns of intra- and inter-modal interactions that underlie the generation of the modality-specific and modality-shared features. This structure enables a detailed examination of how the multimodal features, modality-specific or modality-shared, contribute to survival prediction and which unimodal feature interactions drive them, highlighting the interpretable nature of the disentangled multimodal representations.

\subsection*{Disease-specific survival prediction}
Using the disentangled representations, the model performs disease-specific survival (DSS) prediction and patient risk stratification. We evaluated our approach on four public cancer datasets from The Cancer Genome Atlas (\href{https://www.cancer.gov/ccg/research/genome-sequencing/tcga}{TCGA}): Breast Invasive Carcinoma (BRCA, \textit{n}=928), Bladder Urothelial Carcinoma (BLCA, \textit{n}=423), Lung Adenocarcinoma (LUAD, \textit{n}=463), and Kidney Renal Clear Cell Carcinoma (KIRC, \textit{n}=346).

\subsubsection*{Survival prediction}
DSS was evaluated using the Concordance index (C-index) (Table~\ref{tab:dssresults}) and its inverse probability of censoring–weighted variant (C-index IPCW)~\cite{uno2011c} (Table~\ref{tab:IPCWres}).
Both metrics quantify the agreement between predicted risk scores and observed survival times, with values closer to 1.0 indicating more accurate risk ranking. 

As shown in Table~\ref{tab:dssresults}, both unimodal and multimodal models outperformed the CoxPH model~\cite{cox1972regression} (clinical baseline model using age and sex as covariates), demonstrating the value of leveraging the more complex modalities (\textit{i.e.}, WSIs and transcriptomics).
Among the WSI-based models, prototyping approaches (\textit{i.e.}, DIMAFx$_\text{wsi}$ and PANTHER~\cite{panther}) outperformed ABMIL~\cite{ilse2018attention}, highlighting the effectiveness of summarizing patch-level features into compact representations rather than using all patch features individually.
Transcriptomics-based models showed more uniform performance across variants, with pathway-informed approaches (Pathway MLP and DIMAFx$_\text{tx}$) providing small improvements over the standard MLP. When comparing both modalities, transcriptomics-based models overall outperformed WSI-based models, achieving generally higher C-index values.

Except for PIBD~\cite{pibd}, multimodal models outperformed the unimodal baselines. However, the results vary by dataset: in the smaller BLCA and KIRC cohorts, some unimodal models exceeded certain multimodal baselines. For instance, in BLCA, all transcriptomics-based models outperformed PIBD, SurvPath~\cite{survpath}, MMP~\cite{mmp}, and DIMAFx$_\text{nodis}$. In contrast, in the larger BRCA and LUAD cohorts, most multimodal models consistently performed better than the unimodal baselines.
Among multimodal approaches, prototype-based models (DIMAF, DIMAFx, MMP) outperformed methods that use all or randomly sampled WSI patch features directly (SurvPath, PIBD). This difference is likely in part attributable to the loss function: prototype-based models can be trained with batch sizes greater than one and optimize a Cox proportional hazards loss, whereas PIBD, SurvPath, and ABMIL are restricted to batch sizes of one and thus have to use a negative log-likelihood objective~\cite{mmp}. 
Both our models, DIMAF and DIMAFx, substantially outperformed the other disentanglement baseline, PIBD, and achieved state-of-the-art performance compared to the non-disentanglement approaches SurvPath, MMP, and DIMAFx$_\text{nodis}$. These findings show that the DC loss provides additional regularization without compromising predictive accuracy.

When evaluated using the IPCW-adjusted C-index (Table~\ref{tab:IPCWres}), results were overall lower, reflecting the known upward bias of the standard C-index in heavily censored datasets~\cite{uno2011c}. Interestingly, this effect was not observed in KIRC despite high censoring, potentially reflecting stronger late-event predictability.
The relative ranking of the unimodal models shifted under the IPCW adjustment. While the transcriptomics-based models outperformed WSI-based models using the standard C-index, WSI models showed slightly better performance with IPCW. This is likely due to the WSI representations being more robust as they are derived from UNI~\cite{uni}, a pretrained foundation model. In contrast, the transcriptomics-based models rely on MLPs trained from scratch and thus primarily learn from the observed events in the data. 
Most multimodal approaches still outperformed the unimodal baselines, except for SurvPath and PIBD. Both DIMAF and DIMAFx continued to achieve the best results under IPCW adjustment.

For the BRCA dataset, overall performance decreased substantially, as it exhibited the highest percentage of censored cases. The performance decline was particularly strong for multimodal models compared to unimodal ones, with DIMAFx$_{\text{wsi}}$ achieving the highest IPCW-adjusted C-index, followed by the CoxPH baseline. 
This unexpected result may reflect overfitting of the multimodal models to the (more prevalent) early events, which could lead them to underrepresent the few late events. In contrast, DIMAFx$_{\text{wsi}}$ and the CoxPH model appear more robust to such censoring within this specific dataset.

Overall, these results show that DIMAF and DIMAFx achieve state-of-the-art survival prediction performance. While DIMAF slightly outperformed DIMAFx in terms of C-index, DIMAFx showed improved performance when adjusting for the bias of the C-index (using C-index IPCW), followed by improvements in risk stratification and disentanglement of the multimodal representations, as detailed in the following sections.

\subsubsection*{Risk stratification}
\begin{figure}[H]
    \centering
    \includegraphics[width=\linewidth, trim=0 380 0 0]{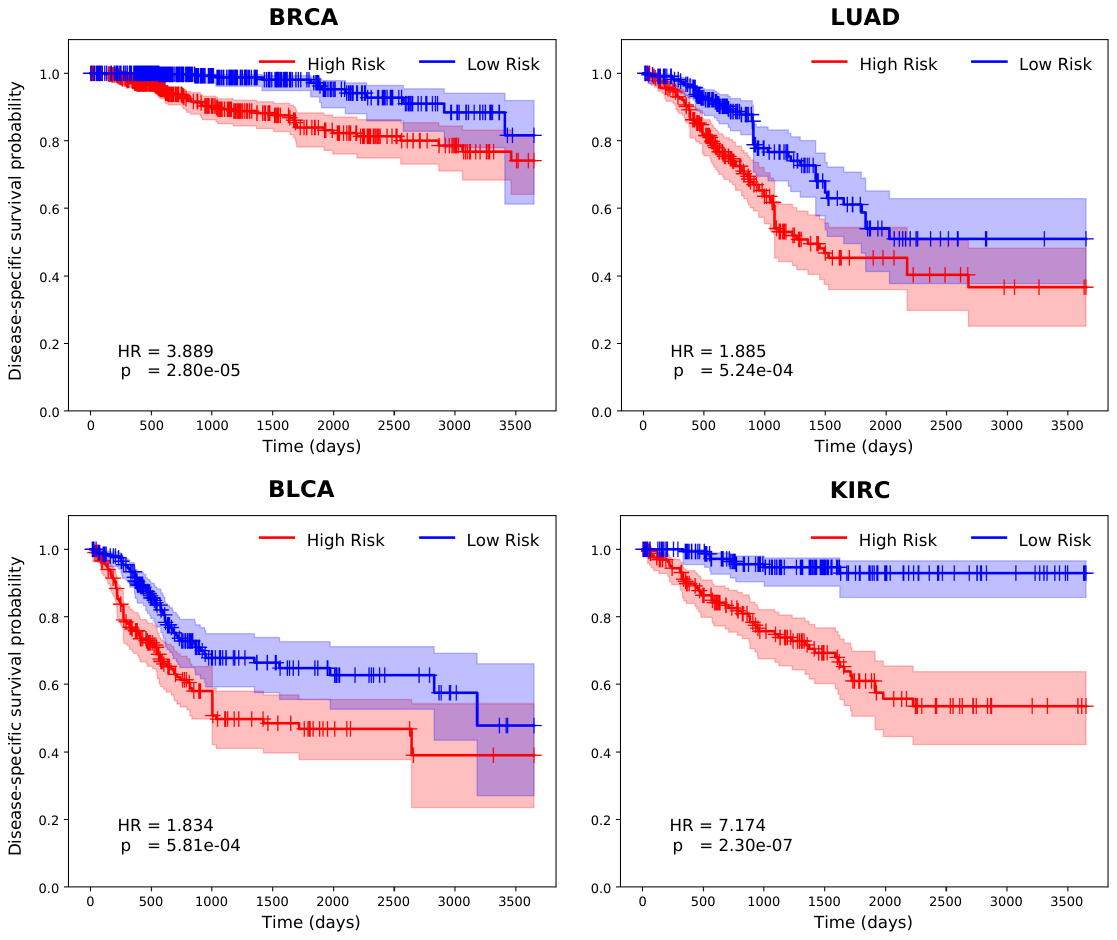}
    \caption{Kaplan–Meier survival curves for DIMAFx, showing high- and low-risk test patient groups stratified by the median predicted risk score, with corresponding hazard ratios and p-values.}
    \label{fig:km_analysis_1}
\end{figure}

We further evaluated our model on risk stratification using Kaplan–Meier survival analysis (Figure~\ref{fig:km_analysis_1}). DIMAFx significantly stratified high- and low-risk patient groups across all four cancer types, achieving Hazard Ratios (HR) above 1 with $p < 0.05$. 
The ablation variant without the disentanglement loss (Figure S1) performed comparably, significantly stratifying the risk groups in all cohorts, with some dataset-specific variability.
When compared with other multimodal baselines, DIMAFx shows similar results as DIMAF (Figure S2) and MMP (Figure S3), both of which consistently produced significant separations across the four datasets. In LUAD, however, DIMAFx produced stronger group separation than the competing methods, with a p-value of $5.24 \cdot 10^{-4}$. SurvPath (Figure S4) achieved significant stratification across all cancer types, but with weaker separations (lower HR, higher p-values). In contrast, PIBD (Figure S5) significantly stratified only two of the four cohorts, namely BLCA and KIRC.

\subsection*{Multimodal disentangled representations}
We evaluated representation disentanglement using Distance Correlation (DC) (Equation~\ref{eq:dcor}) across all cohorts (Table~\ref{tab:DC_res}). DC measures the dependence between two representations, capturing both linear and nonlinear relationships, with DC=0 meaning perfect disentanglement~\cite{empericaldcor,liu2020metrics}. Specifically, we measured disentanglement between the two modality-specific representations (D1), between the modality-specific and modality-shared representations (D2), and as the total disentanglement (\textit{i.e.}, the average of D1 and D2). 
DIMAFx improved over DIMAF, achieving a $10.1\%$ increase in the average total disentanglement. We attribute this improvement to its novel aggregation method ((A) in Figure~\ref{fig:DIMAFx} and described in Section~\ref{sec:daf}) that more effectively combines features within the multimodal disentangled representations. The variant without the disentanglement loss (DIMAFx$_\text{nodis}$) and PIBD both failed to disentangle the multimodal representations, consistent with trends reported in our previous work~\cite{dimaf}.

We further quantified disentanglement using an orthogonal score, defined as the absolute cosine similarity. In contrast to DC, this score measures only linear dependencies. We measured the D1, D2, and total disentanglement across folds (Table S1). Interestingly, DIMAFx$_\text{nodis}$ and DIMAFx showed similar orthogonal scores, both substantially outperforming PIBD, indicating that our model's architecture already achieves linear separation without explicit enforcement. However, nonlinear disentanglement (as measured by DC) requires the disentanglement loss, highlighting its role in promoting nonlinear independence. 

Overall, while DIMAFx and DIMAF both achieve state-of-the-art survival prediction performance, DIMAFx outperforms DIMAF in terms of disentanglement. Both models substantially outperform PIBD (on both counts), demonstrating the effectiveness of the proposed architecture combined with the DC-based loss in disentangling modality-specific and modality-shared representations.

Lastly, we quantified the relative contributions of the modality-specific and modality-shared representations~\cite{dimaf} to verify that all representations are used in survival prediction and to rule out representation collapse, which would also undermine the disentanglement results. As shown in Table S2, both DIMAFx and DIMAF rely mostly on the shared representations, but the modality-specific representations still contribute meaningful information. This highlights the importance of also explicitly modeling modality-specific representations, as they capture complementary signals relevant for survival prediction.

\subsection*{Interpretation of the learned features and their interactions}
To further understand what kind of information the modality-specific and modality-shared representations exactly capture, we conducted an in-depth interpretability analysis on the BRCA dataset. We approached this from both unimodal and multimodal perspectives by addressing three key questions:
\begin{enumerate}
    \item[1.] Which unimodal features contribute most to predicting cancer survival?
    \item[2.] Which multimodal features, modality-shared or modality-specific, drive this prediction? 
    \item[3.] Which unimodal feature interactions are important in forming these multimodal features?
\end{enumerate}

\subsubsection*{Unimodal perspective}
To interpret the WSI-derived prototypical features that were learned by the model, a pathologist reviewed and annotated the prototypes. Table S5 shows these prototype annotations for both the training and test sets together with the prototype cardinality (\textit{i.e.}, the prevalence of the prototype in each set). The prototypes capture distinct morphological patterns such as tumor, stroma, and adipose tissue, with some overlap between prototypes, but also showing variation within a single prototype (\textit{e.g.}, different tumor grades).
The pathway-derived transcriptomic features require no additional annotation, as their predefined pathway labels already provide biological interpretability. We list all pathways in Tables S3 and S4.

\begin{figure}[H]
    \centering
    \includegraphics[width=\textwidth, trim=0 70 0 0]{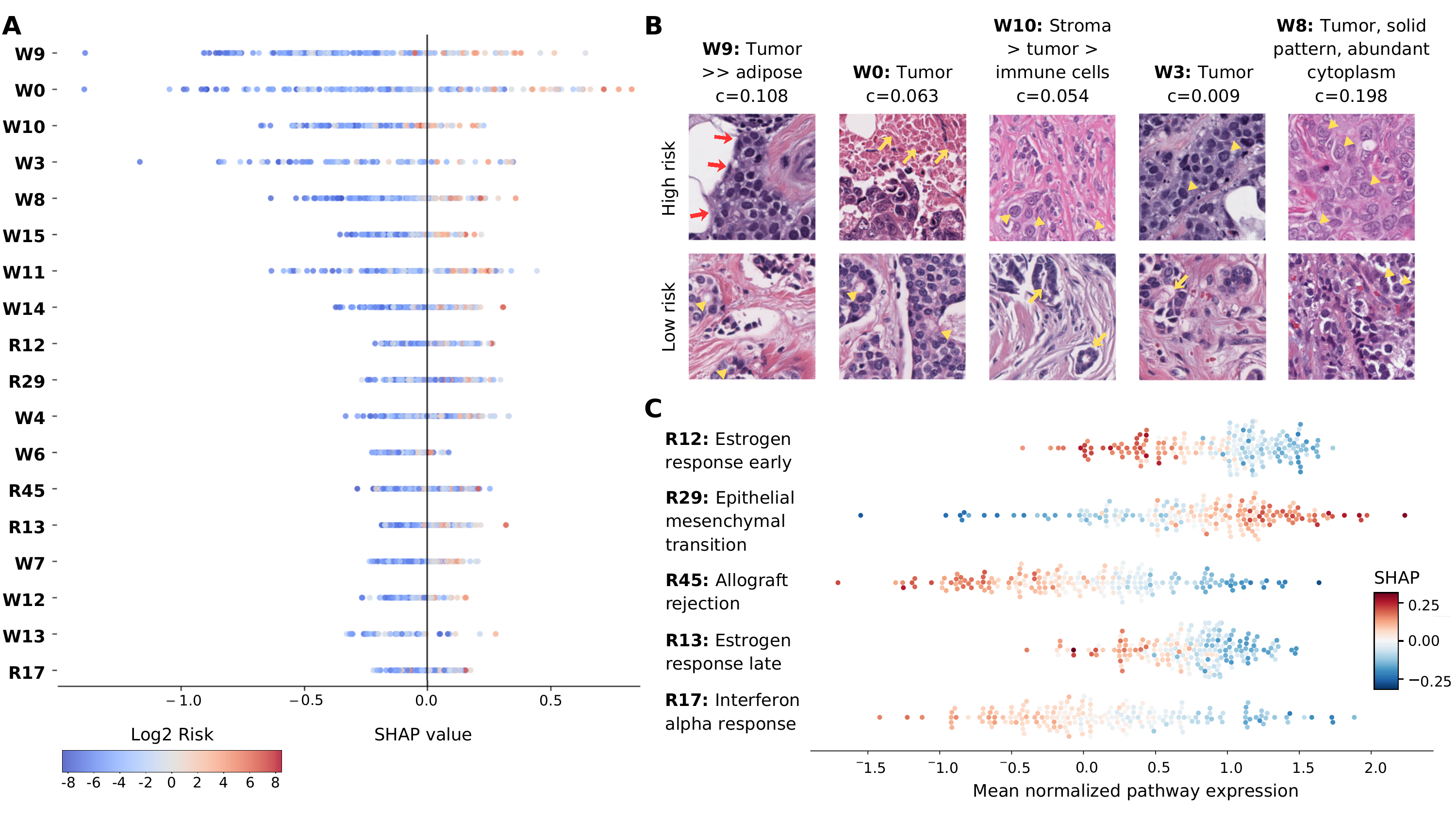}\vspace{0.3cm} 
    \caption{Unimodal interpretability analysis of DIMAFx. \textbf{(A)} SHAP values for the top unimodal features, ranked by mean absolute SHAP value and colored by predicted log2 risk. \textbf{(B)} Visualization of the five most important WSI prototype features. Per feature, the cardinality (c) of the prototype is displayed, indicating their average frequency across all test samples, and two patches representing the prototype are shown, one associated with a high-risk prediction and one with a low-risk prediction. \textbf{W9:} High-risk patch shows solid sheets of tumor cells with dark, hyperchromatic nuclei infiltrating adipocytes (arrows); low-risk patch shows occasional tubule formation (arrowheads) and lower nuclear grade. \textbf{W0}: High-risk patch exhibits solid growth, marked pleomorphism, vesicular chromatin, and tumor necrosis (arrows); low-risk patch shows occasional gland formation (arrowheads), more uniform cells, and lower nuclear atypia. \textbf{W10}: High-risk patch shows solid tumor nests with marked atypia (arrowheads) in stroma containing few immune cells; low-risk patch shows tubular structures (arrows) with low nuclear atypia and sparse immune cells. \textbf{W3}: High-risk patch shows solid tumor nests with basophilic vacuolated cytoplasm, pleomorphic nuclei (arrowheads), and apoptotic cells; low-risk patch shows gland formation (arrow) with lower nuclear atypia. \textbf{W8}: High-risk patch shows solid nests with abundant cytoplasm, vesicular nuclei (arrowheads), and marked atypia; low-risk patch shows solid growth with lower nuclear atypia (arrowheads), suggesting lower histologic grade. \textbf{(C)} Visualization of the five most important transcriptomic pathway features, illustrating how the model-assigned feature risk (SHAP values) varies with the mean log-transformed RSEM-normalized pathway expression.}
    \label{fig:unimodalinterpretation}
\end{figure}

To identify which of these unimodal features are important for survival prediction, we computed SHAP values to assess their contribution to the predicted risk. In Figure~\ref{fig:unimodalinterpretation}A, the calculated SHAP values for the top unimodal features are shown, ordered by the mean absolute SHAP value and colored by the predicted log2 risk.  Unimodal features from WSIs contribute most strongly to the survival prediction, with the top prototypical features comprising mostly tumor morphology (\textit{i.e.}, W9: Tumor $>>$ adipose; W0: Tumor; W10: Stroma $>$ tumor $>$ immune cells; W3: Tumor; W8: Tumor, solid pattern, abundant cytoplasm). From the transcriptomic pathway features, the most important features in relation to the survival prediction are ``R12: Estrogen response early" and ``R29: Epithelial to mesenchymal transition". 

Figure~\ref{fig:unimodalinterpretation}B presents the top five WSI features identified in our SHAP analysis, showing two representative patches for each prototypical feature: one from a high-risk test prediction, and one from a low-risk test prediction. These examples illustrate the variability within each prototype and under which conditions the model associates a prototype feature with a higher or lower risk. Moreover, we show the cardinality (c) for each prototypical feature (\textit{i.e.}, their relative frequency) in the test set.
Across the W0, W8, W9, W10, and W3 features, high-risk patches generally display solid growth patterns, marked nuclear atypia or pleomorphism, vesicular chromatin, or tumor necrosis, features consistent with high-grade morphology. In contrast, low-risk patches exhibit occasional gland or tubule formation and more uniform cells with lower nuclear atypia, reflecting overall lower histologic grades.

Figure~\ref{fig:unimodalinterpretation}C shows the top five pathway features identified in our SHAP analysis, visualizing the relationship between mean pathway expression (log-transformed RSEM-normalized counts) and the SHAP value, which reflects each feature’s contribution to the model-predicted risk.
Across all five features, changes in mean pathway expression are reflected in systematic variation in SHAP values.
For example, both ``R12: Estrogen response early"and ``R13: Estrogen response late" exhibit higher SHAP values at lower pathway expression levels and lower SHAP values at higher expression levels, indicating that reduced pathway expression contributes to higher predicted risk within the model. Estrogen response pathways capture transcriptional processes activated by estrogen receptor (ER) signaling, a key driver of tumor proliferation and progression in many hormone-dependent malignancies, including breast cancer~\cite{hallmark,oh2006estrogen,perou2000molecular,loi2009gene}.
In contrast, higher mean pathway expression of the epithelial-to-mesenchymal transition (EMT) pathway (R29) aligns with increased SHAP values, and thus an increased risk in the model. Elevated activity of the EMT pathway has been shown to promote invasion, metastatic dissemination, and therapy resistance~\cite{yeung2017epithelial,ye2017upholding}, supporting this association between transcriptional pathway activity and clinical outcome.

\subsubsection*{Multimodal perspective}
To identify which multimodal features contribute most strongly to survival prediction, we examined the SHAP values of the modality-shared and modality-specific features. Each unimodal feature is associated with two multimodal features: one in the modality-specific representation and one in the modality-shared representation. The modality-specific feature captures information of the unimodal feature contextualized by information from the same modality and is formed through intra-modal connections in a self-attention layer (see Figure~\ref{fig:DIMAFx}, middle). In contrast, the modality-shared feature integrates information from the other modality via cross-modal interactions in a cross-attention layer.

\begin{figure}[H]
    \centering
    \includegraphics[width=\textwidth, trim=0 630 0 0]{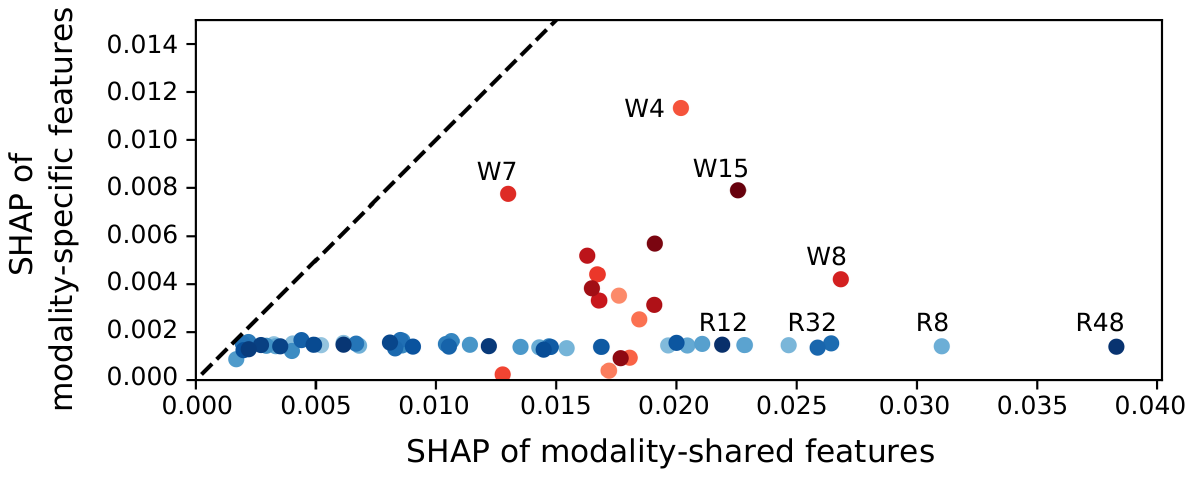}
    \caption{Normalized SHAP values for modality-shared versus modality-specific features showing that shared features generally contribute more to survival prediction. WSI-derived features are colored red, and transcriptomic pathway-derived features are colored blue. The dotted line indicates when the SHAP of the modality-specific feature is equal to the SHAP of the modality-shared feature.}
    \label{fig:multimodalinterpretation_shap}
\end{figure}

In Figure~\ref{fig:multimodalinterpretation_shap}, we compare, for each unimodal feature, the normalized SHAP value of its modality-shared feature with that of its modality-specific counterpart. From the figure, it is evident that the modality-shared features contribute more to survival prediction than modality-specific features, showing that not only the shared information is generally more important (as already shown in Table S2), but also that this holds for each feature individually.
Pathway-derived (blue dots) modality-specific features exhibit uniformly low SHAP values, suggesting that transcriptomic pathways contribute mostly through cross-modal interactions rather than through intra-modal interactions. However, a subset of WSI-derived (red dots) modality-specific features that are related to adipose, stromal, and immune cell morphological characteristics (\textit{i.e.}, W7, W4, and W15) show higher SHAP values. This suggests that certain predictive information in the tumor microenvironment is not encoded by the pathway features. Nonetheless, the most important features are modality-shared features such as ``Shared R48" (R48: KRAS signaling down pathway contextualized by WSI features), ``Shared R8" (R8: G2M checkpoint pathway contextualized by WSI features), ``Shared R32" (R32: Fatty acid metabolism pathway contextualized by WSI features), and ``Shared R12" (R12: Estrogen response early pathway contextualized by WSI features), along with ``Shared W8" (W8: Tumor, solid pattern, abundant cytoplasm prototype contextualized by pathway information).

To further explore these multimodal features and identify which unimodal feature interactions most strongly influence them, we highlight three multimodal features. Specifically, we examine the modality-shared feature derived from W8 (Shared W8), the modality-shared feature derived from R48 (Shared R48), and the modality-specific feature derived from W7 (Specific W7).\\

\begin{figure}[H]
    \centering
    \includegraphics[width=\textwidth, trim=0 240 0 40]{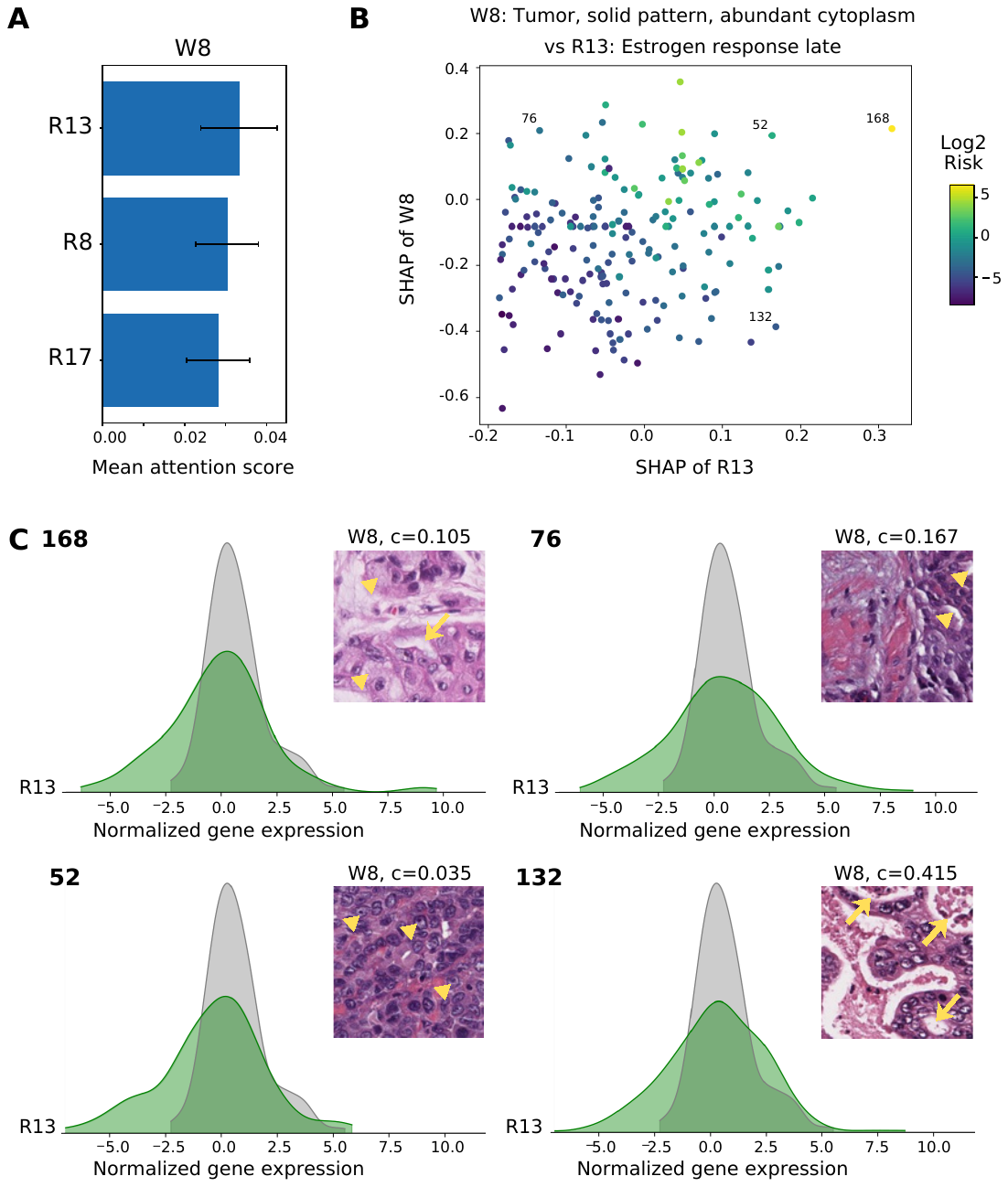}
    \caption{Multimodal interpretability analysis of DIMAFx for the Shared W8 feature, representing a solid-pattern tumor WSI feature contextualized by transcriptomic pathway information. \textbf{(A)} The most highly attended pathway features. \textbf{(B)} Interaction between the W8 and R13 features across all test samples. Each point represents a test case, with SHAP values for R13 (x-axis) and W8 (y-axis) reflecting the model-assigned feature risk, and is colored by the final predicted risk. \textbf{(C)} Visualization of the W8 and R13 features for four representative cases. The ridge plots show the frequency distributions of log transformed RSEM normalized gene expression values for the ``R13: Estrogen response late" pathway in the highlighted case (green) and averaged over all train samples (gray). For the ``W8: Tumor, solid pattern, abundant cytoplasm" feature, we show a representative patch and the cardinality (c) of this morphological prototype from each case. \textbf{Case 168}: the W8 prototype shows a tumor with a solid growth pattern with focal gland formation (arrow), large vacuolated cytoplasm (arrowheads), and high-grade nuclear atypia with conspicuous nucleoli. \textbf{Case 76}: tumor cells forming solid sheets with vague tubule formation (arrowheads) in a collagenous stroma, with scant cytoplasm. \textbf{Case 52}: tumor cells forming solid sheets with vesicular chromatin and conspicuous nucleoli (arrowheads). \textbf{Case 132}: glandular growth pattern (arrows) with more uniform cells and lower nuclear atypia.}
    \label{fig:multimodalinterpretation_W8} 
\end{figure}

\noindent \textbf{Shared W8:} This modality-shared feature is derived by cross-attending the ``W8: Tumor, solid pattern, abundant cytoplasm" prototype feature to all pathway features, yielding a feature of this morphology in the context of pathway information. In Figure~\ref{fig:multimodalinterpretation_W8}A, we show the top three most attended pathway features, ``R13: Estrogen response late", ``R8: G2M checkpoint", and ``R17: Interferon alpha response" with their mean attention values. Figure~\ref{fig:multimodalinterpretation_W8}B illustrates the interaction between W8 and its most attended pathway feature, R13. A clear diagonal color gradient can be seen, highlighting a strong association between this single interaction in the model and the final predicted risk. Although this interaction represents only a small component of the full model, it demonstrates that by first selecting an important multimodal feature and then inspecting its most influential unimodal interactions, we can identify a key unimodal feature interaction that is strongly associated with the final predicted risk.

In Figure~\ref{fig:multimodalinterpretation_W8}C, we look at some representative cases, namely case 168, 76, 52, and 132, to further characterize the interaction between W8 and R13. Case 168 corresponds to a test sample in which both W8 and R13 features contributed strongly to a high risk, \textit{i.e.}, both features have a high SHAP value. In comparison, case 76 shows a similar SHAP value for the W8 feature but a much lower SHAP value for R13. 
Considering the W8 feature alone, both cases have a similar cardinality for this prototype and display morphologies consistent with an intermediate histologic grade. However, inspection of the ridge plots reveals differences in pathway context: the frequency distribution of the normalized gene expression values of the R13 pathway in case 168 is flatter and shifted toward lower values, consistent with this sample being ER negative. In contrast, the R13 pathway distribution in case 76 is not shifted to the left, consistent with this sample having an ER-positive status.
 Through cross-attention, the model integrates the W8 and R13 features into a representation that emphasizes information shared between the two modalities. As a result, similar morphological patterns can contribute differently to the shared multimodal feature depending on differences in the estrogen response late pathway context.
 This difference is reflected in the final predicted risk, with the ER-positive case (76) receiving a substantially lower risk estimate than the ER-negative case (168).

Conversely, cases 52 and 132 have similar SHAP values for the R13 feature, but differ for the W8 feature, which is consistent with the difference in histological grade apparent from their representative patches. The W8 prototype feature from case 52 shows tumor cells forming solid sheets with vesicular chromatin and conspicuous nucleoli, consistent with high-grade morphology. In contrast, the W8 prototype feature from case 132 exhibits a glandular growth pattern with more uniform cells and lower nuclear atypia, with a very high frequency. The R13 pathway distributions of both cases are slightly shifted to the left, but subtle differences between the two distributions remain. 
Although both cases have an ER-negative status, this analysis suggests that small differences in the pathway distributions may interact with the visibly more pronounced differences in morphology in forming the shared feature. This is reflected in the model’s predictions, with case 52 (high-grade morphology) assigned a higher risk than case 132.\\

\begin{figure}[H]
    \centering
    \includegraphics[width=\textwidth, trim=0 330 0 0]{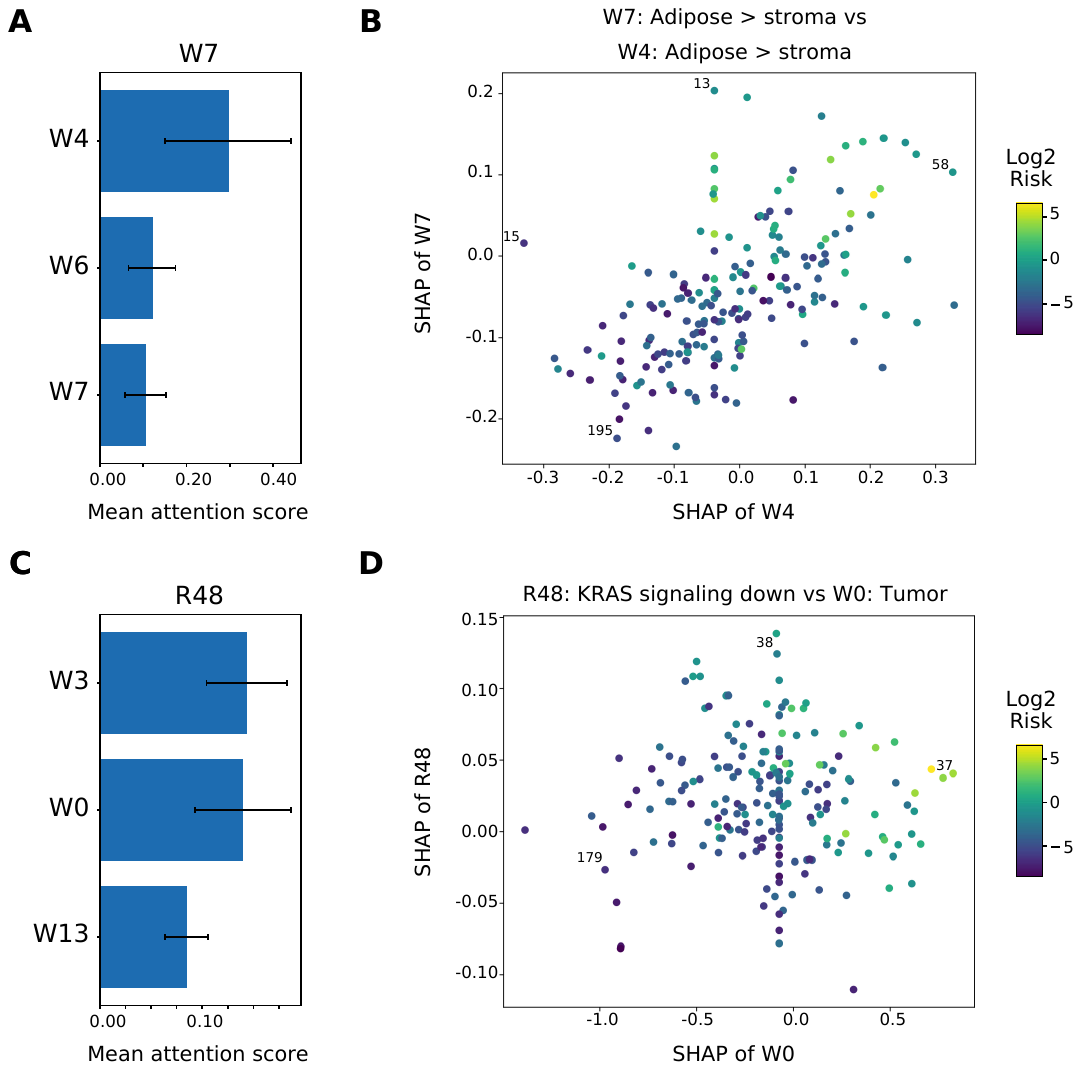}
    \caption{Multimodal interpretability analysis of DIMAFx into two multimodal features. \textbf{(A, B)} ``Specific W7", an adipose/stroma WSI-specific feature; showing \textbf{(A)} the most highly attended WSI features by W7, and \textbf{(B)} the interaction between the ``W7: Adipose $>$ stroma" and ``W4: Adipose $>$ stroma", plotting the SHAP value of the two features. Each sample is colored by the final predicted log2 risk. 
    \textbf{(C, D)} ``Shared R48", KRAS signaling down pathway contextualized by WSI features; 
    showing \textbf{(C)} the most highly attended WSI features, and \textbf{(D)} the interaction between the ``R48: KRAS signaling down" and ``W0: Tumor", plotting the SHAP value of the two features, colored by the final predicted risk.}
    \label{fig:multimodalinterpretation_W7R48} 
\end{figure}

\noindent \textbf{Specific W7:} This modality-specific feature is derived from the adipose tissue with stroma WSI feature (W7) contextualized through self-attention with all WSI features. ``W7: Adipose$>$stroma" attends mostly to other adipose and stromal prototype features as shown in Figure~\ref{fig:multimodalinterpretation_W7R48}A. The interactions between W7 and W4 show a strong diagonal trend between the unimodal SHAP values, suggesting strongly correlated signals related to the predicted risk between both features (Figure~\ref{fig:multimodalinterpretation_W7R48}B). The color gradient is less defined than with the previous interaction, but we can see that overall, the interaction between high-risk features is still associated with a high predicted risk and vice versa.
We further explore this interaction by looking at representative cases 15, 195, 58, and 13. In these cases, both features primarily represent fibrous tissue (stroma), adipose tissue, or fibroadipose tissue (Figure S6). Notably, the patch representing the W7 prototype feature from case 58 contains adipocytes infiltrated by histiocytes, suggestive of fat necrosis, a lesion that may occur following trauma. However, examining these patches alone, without surrounding tissue or architectural context, does not reveal why they would contribute to tumor grade prediction. Additionally, there does not appear to be a clear relationship between the prototype cardinality and the SHAP value or the final predicted risk.
Nevertheless, the model appears to extract predictive information from the combination of W4 and W7 features, even though this signal is not apparent from inspection of the prototype patches or their cardinality.\\

\noindent \textbf{Shared R48:} This modality-shared feature is derived by cross-attending the KRAS signaling down pathway feature to all WSI prototype features. The ``R48: KRAS signaling down" feature has the highest overall attention with WSI prototypes that contain tumor morphological features, \textit{i.e.}, ``W3: Tumor" and ``W0: Tumor" (Figure~\ref{fig:multimodalinterpretation_W7R48}C). When examining the interaction between R48 and W0 in Figure~\ref{fig:multimodalinterpretation_W7R48}D, we do not observe a clear diagonal relationship between the unimodal SHAP values of R48 and the WSI prototype, unlike what was observed previously. However, we do see a distinct color gradient that is visible across the data interactions, especially along the x-axis, reflecting a noticeable association with the predicted risk by the model.
Figure S7 shows representative cases: a high-risk case (37), an intermediate-risk case (38), and a low-risk case (179). Across these examples, no clear, direct relationship is visible between the W0 morphologic prototype, the R48 pathway distribution, and the predicted risk.
From Figure~\ref{fig:multimodalinterpretation_W7R48}, we observed that the model is capturing a signal from the pathway feature that interacts with the W0 prototype in a way relevant to survival prediction. However, this signal is not directly reflected by simple up- or down-regulation of the KRAS pathway or visual morphology. Instead, the model may, for example, be leveraging more subtle information hidden within the genes comprising this pathway. However, what biological information is exactly encoded in these interactions remains an open question at this stage.

\section*{Discussion}

In this work, we presented DIMAFx, an explainable multimodal survival prediction framework that combines WSIs and transcriptomics data using disentangled attention-based fusion. 
A key advantage of DIMAFx is its explainable design, combining inherently interpretable unimodal and multimodal representations with post-hoc feature importance analyses. At the unimodal level, prototypical WSI features and pathway-based transcriptomic features provide clear biological meaning, allowing insights into which morphological patterns and biological functions drive survival predictions. While other methods can also offer unimodal interpretability, DIMAFx extends this to the multimodal domain, revealing underlying tradeoffs and logic between both data modalities. By separating modality-specific and modality-shared representations, the model makes multimodal representations more transparent and enables a comprehensive analysis of how the two modalities interact and which complementary signals are present in the modalities.

Across four cancer cohorts, DIMAFx achieved state-of-the-art performance in survival prediction and patient risk stratification while improving representation disentanglement compared to previous methods. Moreover, the interpretability analysis on the BRCA dataset revealed biologically meaningful insights. WSI features reflected established prognostic factors, including tumor differentiation defined by growth pattern, cellular atypia, and the presence or absence of necrosis. 
From the pathway features, we observed that pathways associated with the tumor-intrinsic biology (\textit{e.g.}, ``R12: Estrogen response early", ``R13: Estrogen response late", and ``R29: Epithelial to mesenchymal transition"), as well as immune or stromal-related processes  (\textit{e.g.},  ``R17: Interferon alpha response" and ``R45: Allograft rejection"), contribute most strongly to breast cancer survival prediction. Disregulation of estrogen response pathways and increased EMT activity have been linked to decreased survival in breast cancer. The Estrogen Response pathways capture transcriptional processes activated by ER signaling, a major driver of tumor proliferation and progression in hormone-dependent malignancies~\cite{hallmark,oh2006estrogen,perou2000molecular,loi2009gene}.
 Decreased expression of ER-regulated gene sets can indicate worse prognosis and poorer response to hormone therapy in ER-positive breast cancer ~\cite{oh2006estrogen}. 
 Similarly, activation of the EMT pathway has been shown to promote invasion, metastatic dissemination, and therapy resistance~\cite{kalluri2009basics,yeung2017epithelial,dongre2019new}.
 In parallel, the contribution of immune-related pathways aligns with the role of tumor-immune interactions influencing clinical outcomes. The interferon alpha response pathway represents type I interferon signaling, which can support anti-tumor immune activation, but chronic or dysregulated activation often results in an immunosuppressed, pro-tumorigenic microenvironment~\cite{benci2016tumor,sistigu2014cancer}. 
Similarly, the allograft rejection pathway covers genes involved in antigen presentation, T-cell activation, and cytotoxic effector responses. High expression of this pathway frequently reflects an inflamed tumor microenvironment with active immune infiltration. This immune-inflamed state has been associated with improved response to immunotherapy, but also indicates ineffective anti-tumor immunity, which correlates with poorer prognosis in certain cases~\cite{rooney2015molecular,thorsson2018immune}. This shows that DIMAFx learns biological patterns from the data that are consistent with well-established mechanisms in breast cancer biology.

From the multimodal interpretability analysis, we observed that the model learned important associations between tumor-related transcriptomic pathway features and WSI patches, including the previously mentioned Estrogen Response late pathway (R13) with the solid tumor prototype W8.  
Notably, lower expression of ER-regulated gene sets has been linked to more aggressive tumor features, such as higher histological grade~\cite{luo2022clinical}.
Additionally, we observed from the SHAP analysis that the most important modality-shared feature was the KRAS signaling down pathway (R48) contextualized by the WSI prototypes W3 and W0, both representing tumor morphologies. However, the precise biological interpretation of the interaction between the KRAS signaling down pathway and the W0 tumor prototype remains open.
Aberrant KRAS signaling is a common feature of many cancers and reflects oncogenic activation of a key signaling pathway involved in tumor development and progression~\cite{klomp2024defining,choucair2025targeting, uniyal2025kras}. 
While there is no established direct link between KRAS signaling and breast cancer survival, the KRAS pathway encompasses downstream effectors such as the PI3K/AKT and MAPK pathways, which are frequently altered in breast cancer~\cite{tokumaru2020kras}. It is therefore possible that the model captured relationships between these downstream effectors and the tumor morphologies represented by the prototypes. Prior studies have also reported subtype-specific associations, with higher KRAS pathway activation linked to improved survival in triple-negative breast cancer~\cite{tokumaru2020kras}, but worse survival in luminal A tumors~\cite{wright2015ras}.
Alternatively, the model may therefore have learned breast cancer subtype-specific characteristics through the combined contribution of morphological features and KRAS signaling pathway activity to survival prediction. 
Finally, among the modality-specific features, morphological features capturing adipose, stromal, and immune cell morphologies were the most important, indicating that certain predictive information from the tumor microenvironment is not encoded in the pathway features. Notably, previous studies suggest that breast adipose tissue plays a major role in breast cancer risk~\cite{wang2012adipose}.
 Overall, this interpretability analysis demonstrates how DIMAFx moves beyond opaque multimodal representations through the disentanglement of modality-specific and shared information, revealing logic and interactions between the two data modalities, and supporting trust, transparency, and potential biomarker discovery.
 
Nevertheless, several limitations should be acknowledged. 
First, the framework relies on predefined pathway sets that partially overlap and do not include all genes. As a result, the information from genes that fall outside these pathways is lost, and conversely, there may be redundant information from the overlapping genes. Moreover, the relevance of these pathways can vary across cancer types. Ideally, pathway selection would be adapted per cancer type, such as the HER2 pathway for breast cancer analyses.
Another limitation concerns the imbalance in the number of unimodal features between WSIs and transcriptomics. This discrepancy may affect how modality-specific and modality-shared features are formed. In future work, selecting a set of pathways that matches the number of WSI features for each cancer type could facilitate more balanced comparisons. 
Moreover, our current analysis remains static and requires familiarity with both the model implementation and programming. Future work will develop interactive visualization dashboards to make the exploration of the features and their interactions even more in-depth, dynamic, and accessible to non-technical experts.
Finally, although DIMAFx substantially improves disentanglement, it does not achieve complete independence between the latent representations. 

Nonetheless, DIMAFx shows that structured, disentangled multimodal representations, coupled with layered interpretability and explainability mechanisms, can deliver strong survival prediction while providing model interpretability and biologically meaningful insights into both modality-specific contributions and cross-modal data interactions.

\section*{Methods}

\subsection*{Data collection and preprocessing}
This study uses four public cancer cohorts from The Cancer Genome Atlas (TCGA): Breast Invasive Carcinoma (BRCA), Bladder Urothelial Carcinoma (BLCA), Lung Adenocarcinoma (LUAD), and Kidney Renal Clear Cell Carcinoma (KIRC).
We only included patients with all three data modalities available: histopathology WSIs, transcriptomics data, and clinical and follow-up records. For patients with multiple diagnostic WSIs (ranging from 1 to 9 per patient), we treated each slide as an independent sample. This resulted in a total of $928$ BRCA, $423$ BLCA, $463$ LUAD, and $346$ KIRC samples.
Each cohort contains data from multiple collection sites and is heavily right-censored, \textit{i.e.}, patient survival times are often only partially observed due to loss of follow-up or the study ending before the event (\textit{i.e.}, death) occurs. The proportion of censored data varies across the datasets: BRCA contains $93.64\%$ right-censored data, BLCA contains $67.85\%$, LUAD contains $71.71\%$, and KIRC contains $83.24\%$ right-censored data.

We downloaded Hematoxylin and Eosin (H\&E)–stained diagnostic WSIs from the GDC data portal (\url{https://portal.gdc.cancer.gov}). All slides have been scanned on Aperio ScanScope systems at 40X or 20X magnification. We uniformly rescaled the images to a spatial resolution of $0.5 \mu m$ per pixel.

From the UCSC Xena database~\cite{goldman2020visualizing}, we downloaded the pan-cancer normalized transcriptomic profiles. These profiles consist of bulk RNA-Sequencing (RNA-Seq) gene expression values normalized across all TCGA cohorts. Specifically, gene expression was measured using the Illumina HiSeq 2000 platform, with transcript abundance quantified as $\text{log}_2(x+1)$ RSEM-normalized counts~\cite{li2011rsem}. For comparability across cohorts, each gene’s expression was further mean-normalized across all TCGA cohorts.

We downloaded clinical and follow-up data (excluding the disease-specific survival (DSS) endpoints) from cBioPortal~\cite{cerami2012cbio,gao2013integrative,de2023analysis}, and incorporated the DSS labels from the curated dataset by Liu~\textit{et al.}~\cite{liu2018integrated}. They defined DSS events as deaths occurring with a tumor present, thereby approximating mortality directly attributable to the disease. Event times were recorded from initial diagnosis to death with tumor. As follow-up durations varied substantially across patients and cohorts, we truncated survival times to 10 years to reduce variability and retain clinical relevance.

\subsection*{Disentangled multimodal survival prediction}
We developed DIMAFx, an interpretable deep learning model that predicts disease-specific cancer survival by integrating histopathology WSIs and bulk RNA-seq data. Building on our prior work~\cite{dimaf}, the model disentangles the multimodal representations, separating modality-specific from modality-shared information. Compared with DIMAF, DIMAFx provides a full explainability framework built around the disentangled prototype-based approach, enabling systematic investigation of how unimodal features interact to form the disentangled multimodal representations and how these relate to patient risk. In addition, the methodological advances in DIMAFx include: (i) improved interpretability of WSI representations (as explained in Section~\ref{sec:feat}), and (ii) an aggregation layer that learns to identify which modality-specific and shared features contribute most to survival prediction (as explained in Section ~\ref{sec:daf}). 

\subsubsection*{Unimodal feature extraction}\label{sec:feat}
\begin{figure}[H]
    \centering
    \includegraphics[width=\textwidth, trim=0 30 0 0]{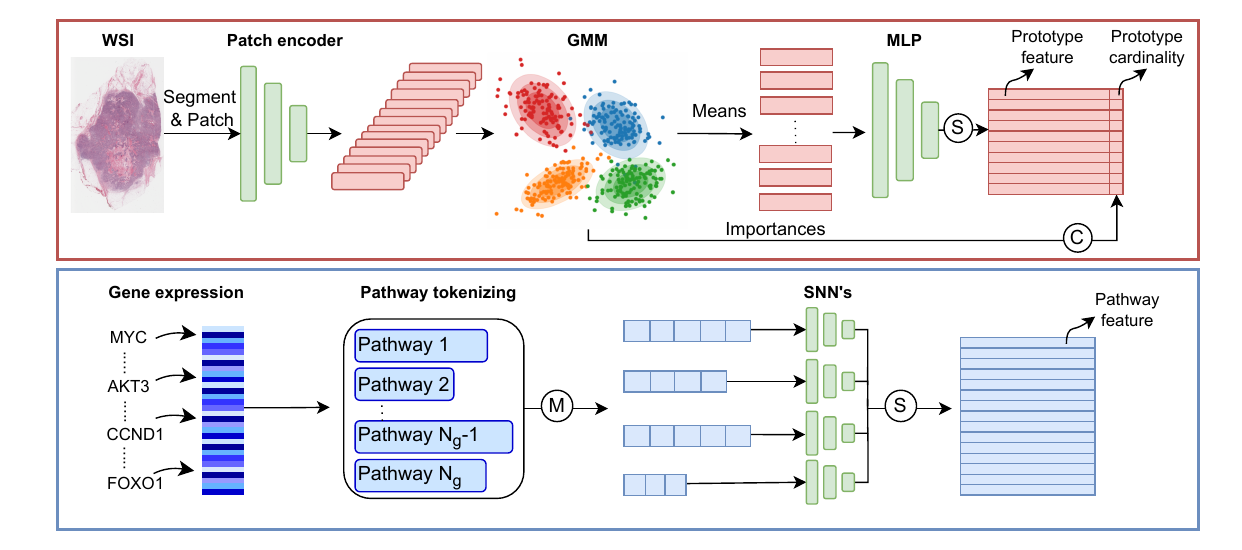}
    \caption{Unimodal feature extraction framework for DIMAFx, for the WSIs (top) and the gene expression data (bottom).}
    \label{fig:unimodalextraction}
\end{figure}

We show the unimodal feature extraction in Figure~\ref{fig:unimodalextraction}. We normalized the WSIs, segmented tissue regions, removed background from each WSI, and divided the images into non-overlapping $256 \times 256$ px patches, using the CLAM framework~\cite{clam}. From these patches, we extracted patch-level features using UNI~\cite{uni}, a DINOv2-based ViT-Large model~\cite{dosovitskiy2020vit,oquab2023dinov2} pretrained on more than $1 \times 10^5$ H\&E-stained WSIs across 20 major tissue types from Mass General Brigham.
Inspired by PANTHER~\cite{panther,mmp}, a prototyping-based approach that leverages the morphological redundancy in WSIs, we summarize the patch features of each slide into a compact and interpretable representation. Specifically, we set the distribution of patch features from each WSI using a Gaussian Mixture Model (GMM) with $N_h = 16$ components, which is optimized via differentiable Expectation-Maximization (EM)~\cite{dempster1977maximum,kim2021differentiable}. Component means are initialized as the centroids obtained from k-means ($k=N_h$) clustering on $160000$ randomly sampled training patches. Covariances are initialized as identity matrices ($I_{N_h}$) and mixture weights as $1/N_h$. After optimization, the learned parameters of the components (\textit{i.e.}, the weights, means, and covariances of the mixture distributions) serve as a compact representation of the slides' content~\cite{panther}. From these parameters, we use only the mean of each Gaussian distribution, representing a morphological prototype, and its corresponding mixture weight, reflecting prototype cardinality, \textit{i.e.}, the proportion of patches in the WSI associated with that prototype. By taking the patches with the highest posterior probability for each Gaussian component, we can visualize these prototypes and identify their dominant morphological patterns (\textit{e.g.}, tumor, stroma, adipose tissue). This illustrates the interpretability of the learned prototypical features. In contrast to our prior work~\cite{dimaf}, the distribution means were passed through a linear layer for dimensionality reduction before being concatenated with their respective mixture weights. This design helps DIMAFx to preserve these important, single-scalar values during the dimensionality reduction process, enabling the model to capture both the type of morphological pattern (prototype) and its frequency within a slide (cardinality). 
Finally, learnable prototype tokens are added to these interpretable WSI features~\cite{mmp} and stacked to form the WSI representation $\mathbf{Z_{h}} \in \mathbb{R}^{N_h\times D}$. 

We mapped the bulk gene expression profiles to $N_g=50$ biological curated pathways~\cite{survpath,mmp,pibd} from the Molecular Signatures Database (MSigDB) hallmark gene set collection~\cite{hallmark} (Table S3 and S4). Each pathway represents a specific biological state or process in cancer~\cite{hallmark} and contains a set of genes whose expression levels characterize that process. For each pathway, the expression values of its member genes were concatenated to form a pathway-level input and processed separately through individual Self-Normalizing Networks (SNNs)~\cite{snn} to create pathway-level features. Similar to the WSI representations, learnable pathway tokens are concatenated to each output, which are then stacked to create the transcriptomics representation  $\mathbf{Z_{g}} \in \mathbb{R}^{N_g\times D}$. This design enables direct interpretability, as each feature encodes a well-defined biological process. 

\subsubsection*{Disentangled Attention Fusion}
\label{sec:daf}
Given the unimodal representations $\mathbf{Z_{h}}$ and $\mathbf{Z_{g}}$, we distinctly model four types of interactions to obtain separate modality-specific and modality-shared representations~\cite{dimaf}. Two self-attention layers model the intra-modal interactions, encoding the information specific to each modality in separate representations, $\mathbf{Z^p_{gg}} \in \mathbb{R}^{N_g \times D_z}$ and $\mathbf{Z^p_{hh}} \in \mathbb{R}^{N_h \times D_z}$. Contrarily, two cross-attention layers model the inter-modal interactions to encode the information that is shared between the modalities in two representations, $\mathbf{Z^p_{hg}} \in \mathbb{R}^{N_g \times D_z}$ and $\mathbf{Z^p_{gh}} \in \mathbb{R}^{N_h \times D_z}$.
In other words, each unimodal feature generates two multimodal features: one contextualized by the information from the same modality (modality-specific) and one by the information from the other modality (modality-shared). This disentangled design enables the model to separate modality-specific signals from shared, cross-modal signals. 

These multimodal features in the four obtained representations ($\mathbf{Z^p_{gg}}$, $\mathbf{Z^p_{hh}}$, $\mathbf{Z^p_{hg}}$, $\mathbf{Z^p_{gh}}$) are then pooled via an attention-based aggregation layer, adapted from attention-based multiple instance learning~\cite{ilse2018attention}. 
This layer computes attention weights for each feature by passing them through a linear layer followed by a softmax function. The final disentangled representations $\mathbf{z_{gg}}$, $\mathbf{z_{hh}}$, $\mathbf{z_{hg}}$, $\mathbf{z_{gh}} \in \mathbb{R}^{D_z}$, are then obtained by computing a weighted mean of the features within each representation based on these attention weights. This mechanism enables DIMAFx to identify, for each patient, which features in each latent space are most informative for predicting the risk.

To further encourage these representations to capture distinct signals, we use a disentanglement loss based on Distance Correlation (DC)~\cite{empericaldcor,liu2020metrics}. 

\begin{equation}\label{eq:dcor}
    \text{DC}(\mathbf{Z_1, Z_2}) = \frac{dCov(\mathbf{Z_1, Z_2})}{\sqrt{dCov(\mathbf{Z_1, Z_1})dCov(\mathbf{Z_2, Z_2})}}
\end{equation}

\noindent Here, $dCov(\mathbf{X, Y})$ denotes the (Euclidean) distance covariance between $\mathbf{X}$ and $\mathbf{Y}$~\cite{empericaldcor}. It measures the dependence between $\mathbf{X}$ and $\mathbf{Y}$ by assessing how variations in pairwise distances within one correspond to variations in the other. Intuitively, DC is small when two representations vary independently across patients and large when they share dependence. We penalize the dependence between the two modality-specific representations and the dependence between the modality-specific and modality-shared representations. Here, $\mathbf{Z_{gg}, Z_{hh}, Z_{hg}, \text{ and } Z_{gh}} \in \mathbb{R}^{B\times D_z}$ are the $B$ stacked disentangled representations. 

\begin{equation}\label{eq:dcorloss}
    \mathcal{L}_{dis} = DC(\mathbf{Z_{gg}, Z_{hh}}) + DC([\mathbf{Z_{gg}, Z_{hh}}], [\mathbf{Z_{hg}, Z_{gh}}])
\end{equation}

\subsubsection*{Risk prediction}
Following the Cox proportional hazards model~\cite{cox1972regression}, we obtain a risk score $r$ using the concatenated disentangled representations and a linear predictor. This risk score is optimized with the Cox partial log likelihood~\cite{wong1986theory}
\begin{equation}
    \mathcal{L}_{surv} = -\sum_{i \in U}\bigg(r_i - log \big(\sum_{j \in K_i}e^{r_j}\big)\Bigg) \,
\end{equation}
\noindent where $U$ is the set of uncensored patients and $K_i$ is the set of patients whose time of death or last known follow-up time is after $i$. \\

\noindent The overall objective is 
\begin{equation}
    \mathcal{L} = \lambda_{surv} \cdot \mathcal{L}_{surv} +  \lambda_{dis} \cdot \mathcal{L}_{dis}
\end{equation}
\noindent with $\lambda_{\text{surv}} = 1$ and $\lambda_{\text{dis}} = 7$. 

\subsection*{Survival Analysis}
We trained the models for each cancer type to predict DSS risk using 5-fold cross-validation. We stratified the train-test splits by collection site to mitigate potential batch effects~\cite{howard2021impact,mmp}. We trained the models for 30 epochs using the AdamW optimizer with a batch size of $64$, a learning rate of $10^{-4}$, a cosine learning scheduler, and $1e^{-5}$ weight decay. 

We evaluated the model's survival prediction performance using Harrell's Concordance index (C-index) and its weighted variant C-index IPCW (Inverse Probability Censoring Weighting)~\cite{uno2011c}. The C-index measures rank correlation between predicted risk scores and observed time points. However, high censoring can bias the C-index upward~\cite{uno2011c}. To address this, C-index IPCW weighs each case by the inverse probability of censoring to reduce the influence of early censored observations, which inherently carry more uncertainty. The value for both metrics ranges from 0 to 1, with 1 indicating perfect prediction. 

To further evaluate the model's performance, we stratified patients into high- and low-risk groups based on the median predicted risk score. We then used Kaplan–Meier analysis to generate survival curves for each group, and assessed the differences using the log-rank test. We considered p-values below 0.05 statistically significant. We also calculated the hazard ratio (HR) to quantify the relative risk between the two groups. An HR greater than 1 indicates that patients in the high-risk group have a higher risk of the event (\textit{i.e.}, death) compared to those in the low-risk group, thereby confirming that higher predicted risk scores align with worse observed clinical outcomes.

\subsection*{Evaluation of Disentanglement}
We assessed representation disentanglement using Distance Correlation~\cite{empericaldcor,liu2020metrics} (Equation~\ref{eq:dcor}) and an orthogonal score. Both metrics are easy to compute and do not require additional networks (\textit{e.g.}, mutual information estimators). Distance Correlation captures both linear and non-linear dependencies, whereas the orthogonal score is restricted to linear relationships. The orthogonal score is defined as the absolute cosine similarity between vector representations. Both metrics yield values between 0 and 1, with lower values indicating higher disentanglement.  
We evaluated disentanglement across two levels: D1, which quantifies the disentanglement between the two modality-specific representations, and D2, which measures the disentanglement between the modality-specific and the modality-shared representations. The total disentanglement score is reported as the average of D1 and D2.

While DC and the orthogonal score quantify statistical independence, they do not guarantee that a representation is actually leveraged by the predictive model.
Therefore, to assess the contribution of the disentangled representations, we employed deepSHAP~\cite{NIPS2017_7062}, an extension of SHAP (SHapley Additive exPlanations) tailored for deep neural networks. 
SHAP values, derived from cooperative game theory, quantify each feature’s marginal contribution to a prediction and provide several theoretical properties, including local accuracy, consistency, and additivity. 
To make computation tractable, Deep SHAP leverages the DeepLIFT algorithm\cite{shrikumar2017learning} and approximates conditional expectations using background samples. We computed normalized SHAP values for the different disentangled representations after fusion for each of the test samples. To obtain global feature importance, we averaged the absolute SHAP values across all these samples, thereby quantifying the overall contribution of each representation to model predictions.

\subsection*{Biological interpretation}
DIMAFx enables a comprehensive interpretability analysis, providing insights from both unimodal and multimodal perspectives.
We assessed the interpretability of our framework on the TCGA-BRCA dataset using the trained model and the test set of fold 2. 

For the transcriptomics modality, the features are directly interpretable, as each feature represents a single pathway from the MSigDB Hallmark gene set collection~\cite{hallmark}. These pathways consist of genes that are curated and annotated by experts, providing a direct link to their known biological functions. The full list of pathway annotations is provided in Tables S3 and S4.
For the WSI modality, interpretability arises from visualizing and annotating the learned prototype features. By visualizing the patches with the highest posterior probability for each Gaussian component, we obtain representative morphological patterns for each prototype. A pathologist annotated these prototypes (Table S5), identifying distinct morphological structures such as tumor, stroma, and adipose tissue. 

To evaluate the contribution of the individual unimodal features to survival prediction, we applied DeepSHAP to compute the SHAP values for both the WSI and transcriptomics features ($\{\mathbf{z}_{h,i}\}^{N_h}_{i=1}$ and $\{\mathbf{z}_{g,i}\}^{N_g}_{i=1}$) for each test sample. These SHAP values quantify each feature’s contribution to the predicted survival, indicating whether a feature increases (SHAP $>$ 0) or decreases (SHAP $<$ 0) the predicted risk. 
For the transcriptomic features, we further compared their SHAP values with the mean log-transformed RSEM-normalized pathway expression to determine whether (partly) up- or down-regulated pathways are associated with higher or lower predicted risk.
For each WSI feature, we compared instances (patches from samples) with high versus low SHAP values to assess how the associated morphological patterns drive survival predictions. In other words, we examined which morphological patterns are characteristic of a prototype feature when it is given a high- or low-risk.

From the multimodal perspective, we applied DeepSHAP to estimate the SHAP values of the modality-specific features, $\{\mathbf{z}^p_{hh,i}\}^{N_h}_{i=1}$ and $\{\mathbf{z}^p_{gg,i}\}^{N_g}_{i=1}$ and the modality-shared features $\{\mathbf{z}^p_{gh,i}\}^{N_h}_{i=1}$ and $\{\mathbf{z}^p_{hg,i}\}^{N_g}_{i=1}$. This allowed us to identify which multimodal features most strongly influence survival prediction. To further understand how these key multimodal features are formed, we analyzed the attention weights and the contributing unimodal features to identify which unimodal feature interactions most strongly influence their formation and the biological patterns underlying their contribution.

\subsection*{Resource availability}

\subsubsection*{Lead contact}
Requests for further information and resources should be directed to and will be fulfilled by the lead contact, Wilson Silva (w.j.dossantossilva@uu.nl).

\subsubsection*{Materials availability}
This study did not generate new unique reagents.

\subsubsection*{Data and code availability}
\begin{itemize}
    \item All data used in this study are publicly available from The Cancer Genome Atlas (TCGA) database. Specifically, we obtained WSIs directly from \url{https://portal.gdc.cancer.gov}, transcriptomics profiles from \url{https://xenabrowser.net/datapages/}, and clinical data from \url{https://www.cbioportal.org} and \url{https://pmc.ncbi.nlm.nih.gov/articles/PMC6066282/#_ad93_}. Complete instructions, including data access, preprocessing steps, and dataset splits used in this study are publicly available at \url{https://github.com/Trustworthy-AI-UU-NKI/DIMAFx}.
    \item All sources required to reproduce our results are publicly accessible. The source code is publicly available at \url{https://github.com/Trustworthy-AI-UU-NKI/DIMAFx} along wmaith all hyperparameter configurations, a complete list of dependencies, and detailed instructions. 
\end{itemize}

\subsection*{Acknowledgments}
We thank The Cancer Genome Atlas (TCGA) Research Network for making the datasets used in this study publicly available.
This research is part of the project ``Ordinality informed Federated Learning for Robust and Explainable Radiology AI" with file number NGF.1609.241.009 of the research program AiNED XS Europa, which is (partly) financed by the Dutch Research Council (NWO).

\subsection*{Author contributions}
Concept \& design: A.E., W.S.;
Data acquisition, model development \& analysis: A.E.;
Clinical expertise: M.E.C.;
Data interpretation \& biological interpretation of results: S.L., M.E.C., S.P.O.;
Visualization: A.E., S.L., A.C., S.A., W.S.;
Supervision: S.A., W.S.;
Writing - original draft: A.E., S.L., M.E.C.;
Writing - Review \& editing: S.P.O., A.C., S.A., W.S.;
All authors read and approved the final manuscript.

\subsection*{Declaration of interests}
The authors declare no competing interests.

\subsection*{Declaration of generative AI and AI-assisted technologies in the writing process}
During the preparation of this work, the authors used ChatGPT and Grammarly to assist with grammar corrections and improve the readability of the text. The authors have carefully reviewed and edited the content as needed and take full responsibility for the content of the manuscript. 

\section*{Figures, figure titles, and figure legends}

\section*{Tables, table titles, and table legends}

\begin{table}[H]
\centering
\caption{DSS test results measured by C-index (mean $\pm$ std) for each TCGA cancer cohort and the overall average (Avg). The best and second-best performances are denoted by \textbf{bold} and \underline{underlined}, respectively.}
\begin{tabular}{ll|l|llll}
\toprule
\textbf{Modality} & \textbf{Model} & \textbf{Avg.} & \textbf{BRCA}  & \textbf{BLCA} & \textbf{LUAD} & \textbf{KIRC} \\
\midrule
\textbf{Clinical} & CoxPH~\cite{cox1972regression}  & 0.496 & 0.491±0.113 & 0.528±0.073 & 0.463±0.067 & 0.502±0.111 \\
\midrule
\textbf{WSI} & ABMIL~\cite{ilse2018attention} & 0.606 & 0.624±0.083 & 0.541±0.077 & 0.570±0.043 & 0.688±0.135 \\
& PANTHER~\cite{panther} & 0.650 & 0.627±0.071 & 0.605±0.066 & 0.604±0.016 & \textbf{0.763±0.138} \\
& DIMAFx$_{\text{wsi}}$ & 0.653 & 0.699±0.048 & 0.581±0.064 & 0.600±0.024 & 0.733±0.124 \\
\midrule
\textbf{Transcript}  & MLP & 0.654 & 0.692±0.078 & \underline{0.672±0.059} & 0.603±0.049 & 0.650±0.148 \\
\textbf{omics} & Pathway MLP & 0.673 & 0.685±0.045 & 0.663±0.054 & 0.630±0.068 & 0.712±0.075 \\
& DIMAFx$_{\text{tx}}$ & 0.666 & 0.675±0.055 & 0.669±0.059 & 0.612±0.069 & 0.709±0.081 \\ 
\midrule
\textbf{Multimodal} 
 & SurvPath~\cite{survpath} & 0.667 & 0.713±0.027 & 0.610±0.054 & 0.608±0.050 & 0.737±0.100 \\
& MMP~\cite{mmp} & 0.700 & 0.741±0.056 & 0.657±0.040 & \textbf{0.674±0.038} & 0.728±0.106 \\
& DIMAFx$_{\text{nodis}}$ & 0.700 & 0.749±0.054 & 0.660±0.044 & 0.651±0.035 & 0.740±0.117 \\
\cmidrule(lr){2-7}
& PIBD~\cite{pibd} & 0.613 & 0.608±0.100 & 0.570±0.073 & 0.539±0.051 & 0.735±0.101 \\
& DIMAF~\cite{dimaf} & \textbf{0.715} & \underline{0.759±0.065} & \textbf{0.679±0.043} & \underline{0.669±0.062} & \underline{0.752±0.092} \\

& DIMAFx & \underline{0.708} & \textbf{0.760±0.057} & 0.670±0.045 & 0.656±0.051 & 0.745±0.093 \\
\bottomrule
\end{tabular}

\label{tab:dssresults}
\end{table}

\begin{table}[H]
\caption{DSS test results measured by C-index IPCW (mean $\pm$ std) for each TCGA cancer cohort and overall average (Avg). The best and second-best performances are indicated by \textbf{bold} and \underline{underlined}, respectively.}\label{tab:IPCWres}%
\begin{tabular}{ll|l|llll}
\toprule
\textbf{Modality} & \textbf{Model} & \textbf{Avg.} & \textbf{BRCA}  & \textbf{BLCA} & \textbf{LUAD} & \textbf{KIRC} \\
\midrule
\textbf{Clinical} & CoxPH~\cite{cox1972regression} & 0.562 & \underline{0.611±0.192} & 0.576±0.111 & 0.551±0.089 & 0.509±0.116 \\
\midrule
\textbf{WSI} & ABMIL~\cite{ilse2018attention} & 0.593 & 0.593±0.144 & 0.528±0.062 & 0.554±0.061 & 0.698±0.113 \\
& PANTHER~\cite{panther} & 0.618 & 0.543±0.155 & 0.573±0.089 & 0.586±0.038 & 0.771±0.128 \\
& DIMAFx$_{\text{wsi}}$ & 0.619 & \textbf{0.627±0.095} & 0.526±0.088 & 0.588±0.040 & 0.736±0.115 \\
\midrule
\textbf{Transcript} & MLP & 0.576 & 0.559±0.073 & 0.561±0.086 & 0.496±0.118 & 0.686±0.113 \\ 
\textbf{omics} & Pathway MLP & 0.590 & 0.533±0.088 & 0.576±0.114 & 0.494±0.121 & 0.755±0.073 \\ 
& DIMAFx$_{\text{tx}}$ & 0.592 & 0.505±0.139 & \underline{0.609±0.128} & 0.495±0.116 & 0.757±0.073 \\ 
\midrule
\textbf{Multimodal} & SurvPath~\cite{survpath} & 0.603 & 0.551±0.111 & 0.556±0.124 & 0.535±0.080 & 0.770±0.105 \\
& MMP~\cite{mmp} & 0.621 & 0.519±0.185 & 0.606±0.080 & \underline{0.604±0.052} & 0.755±0.102 \\

& DIMAFx$_{\text{nodis}}$ & 0.626 & 0.556±0.142 & 0.598±0.116 & 0.578±0.059 & \underline{0.772±0.101} \\
\cmidrule(lr){2-7} & PIBD~\cite{pibd} & 0.591 & 0.497±0.137 & 0.578±0.070 & 0.524±0.101 & 0.765±0.081 \\
& DIMAF~\cite{dimaf} & \underline{0.635} & 0.580±0.161 & 0.576±0.153 & \underline{0.604±0.074} & \textbf{0.778±0.093} \\
& DIMAFx & \textbf{0.641} & 0.553±0.150 & \textbf{0.610±0.123} & \textbf{0.623±0.091} & \textbf{0.778±0.084} \\
\bottomrule
\end{tabular}
\label{tab:IPCWres}
\end{table}

\begin{table}[H]
\caption{Disentanglement test results measured by DC over the different folds (mean $\pm$ std). We denote the best and second-best total disentanglement by \textbf{bold} and \underline{underlined}, respectively.}\label{tab2}%
\begin{tabular}{lllllll}
\toprule
& & \textbf{BRCA}  & \textbf{BLCA} & \textbf{LUAD} & \textbf{KIRC} & \textbf{Avg. }\\
\midrule
PIBD~\cite{pibd} & D1 & 0.324±0.022 & 0.444±0.010 & 0.444±0.033 & 0.531±0.034 & 0.436 \\
 & D2 &0.924±0.027 & 0.898±0.076 & 0.960±0.019 & 0.930±0.023 & 0.928\\
  & Total &0.624±0.021 & 0.671±0.037 & 0.702±0.022 & 0.731±0.012 & 0.682\\
  \midrule
DIMAF~\cite{dimaf}& D1 & 0.244±0.043 & 0.401±0.029 & 0.339±0.046 & 0.525±0.089 & 0.377 \\
 & D2 & 0.469±0.050 & 0.672±0.068 & 0.609±0.029 & 0.931±0.031 & 0.670\\
  & Total & \underline{0.356±0.045} & \underline{0.537±0.048} & \underline{0.474±0.018} & \underline{0.728±0.053} & \underline{0.524}\\
    \midrule
DIMAFx$_{\text{nodis}}$ & D1 & 0.476±0.023 & 0.619±0.062 & 0.567±0.078 & 0.622±0.082 & 0.571\\
 & D2 & 0.946±0.020 & 0.965±0.017 & 0.955±0.031 & 0.976±0.009 & 0.961 \\
  & Total & 0.711±0.019 & 0.792±0.037 & 0.761±0.051 & 0.799±0.044 & 0.766\\
  \midrule
DIMAFx & D1 & 0.214±0.026 & 0.373±0.062 & 0.298±0.053 & 0.438±0.062 & 0.331\\
 & D2 &0.453±0.080 & 0.631±0.082 & 0.571±0.040 & 0.793±0.159 & 0.612 \\
  & Total & \textbf{0.333±0.041} & \textbf{0.502±0.067} & \textbf{0.435±0.026} & \textbf{0.615±0.107} & \textbf{0.471}\\
\bottomrule
\end{tabular}
\label{tab:DC_res}
\end{table}

\newpage
\bibliography{references}

\newpage
\section*{Supplemental Information}

\setcounter{figure}{0}
\renewcommand{\thefigure}{S\arabic{figure}} 

\setcounter{table}{0}
\renewcommand{\thetable}{S\arabic{table}} 

\begin{figure}[H]
    \centering
        \includegraphics[width=\linewidth, trim=0 380 0 0]{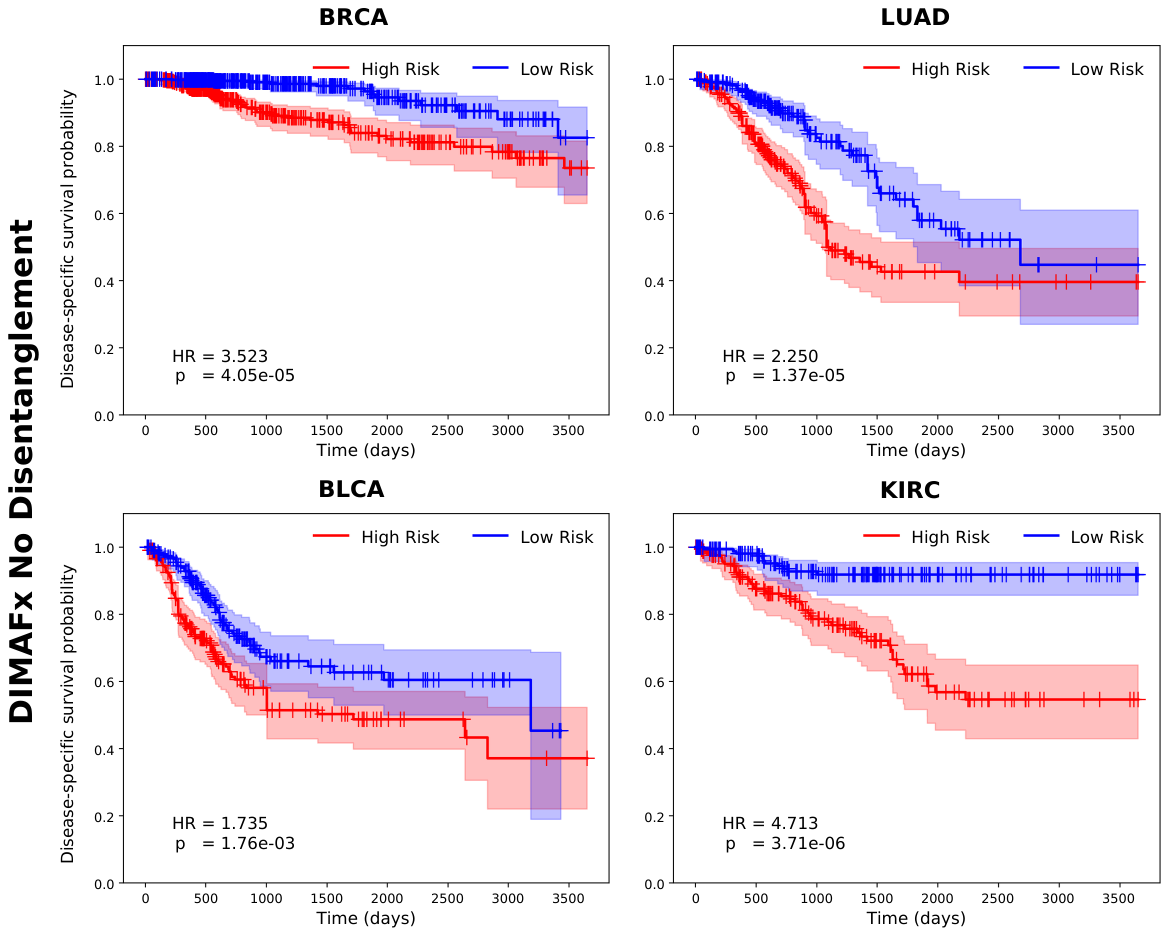}
    \caption{Kaplan–Meier survival curves for DIMAFx$_\text{nodis}$, showing high- and low-risk patient groups stratified by the median predicted risk score, with corresponding hazard ratios and p-values.}
    \label{fig:km_analysis_DIMAFxnodis}
\end{figure}

\begin{figure}[H]
    \centering
        \includegraphics[width=\linewidth, trim=0 380 0 0]{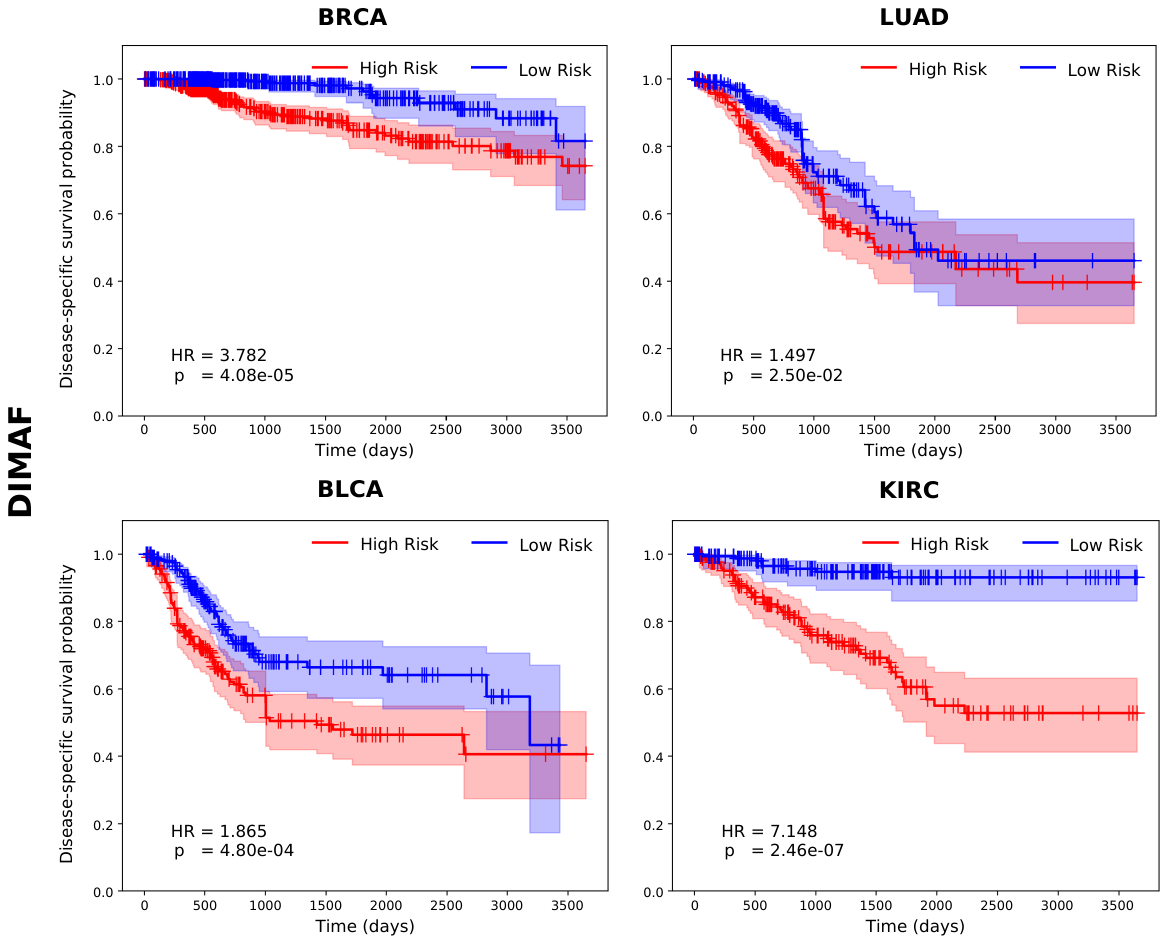}
    \caption{Kaplan–Meier survival curves for DIMAF, showing high- and low-risk patient groups stratified by the median predicted risk score, with corresponding hazard ratios and p-values.}
    \label{fig:km_analysis_dimaf}
\end{figure}

\begin{figure}[H]
    \centering
        \includegraphics[width=\linewidth, trim=0 380 0 0]{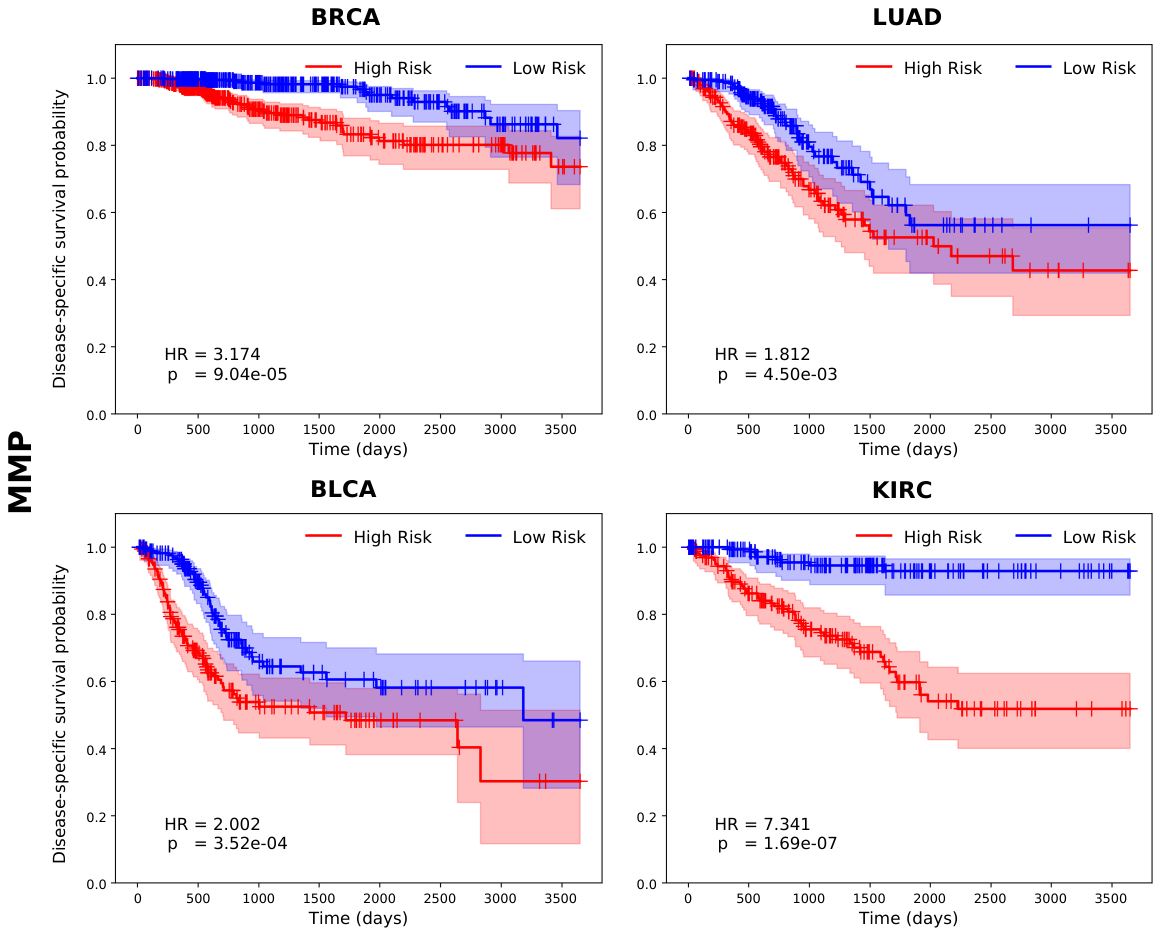}
    \caption{Kaplan–Meier survival curves for MMP, showing high- and low-risk patient groups stratified by the median predicted risk score, with corresponding hazard ratios and p-values.}
    \label{fig:km_analysis_mmp}
\end{figure}

\begin{figure}[H]
        \centering
        \includegraphics[width=\linewidth, trim=0 380 0 0]{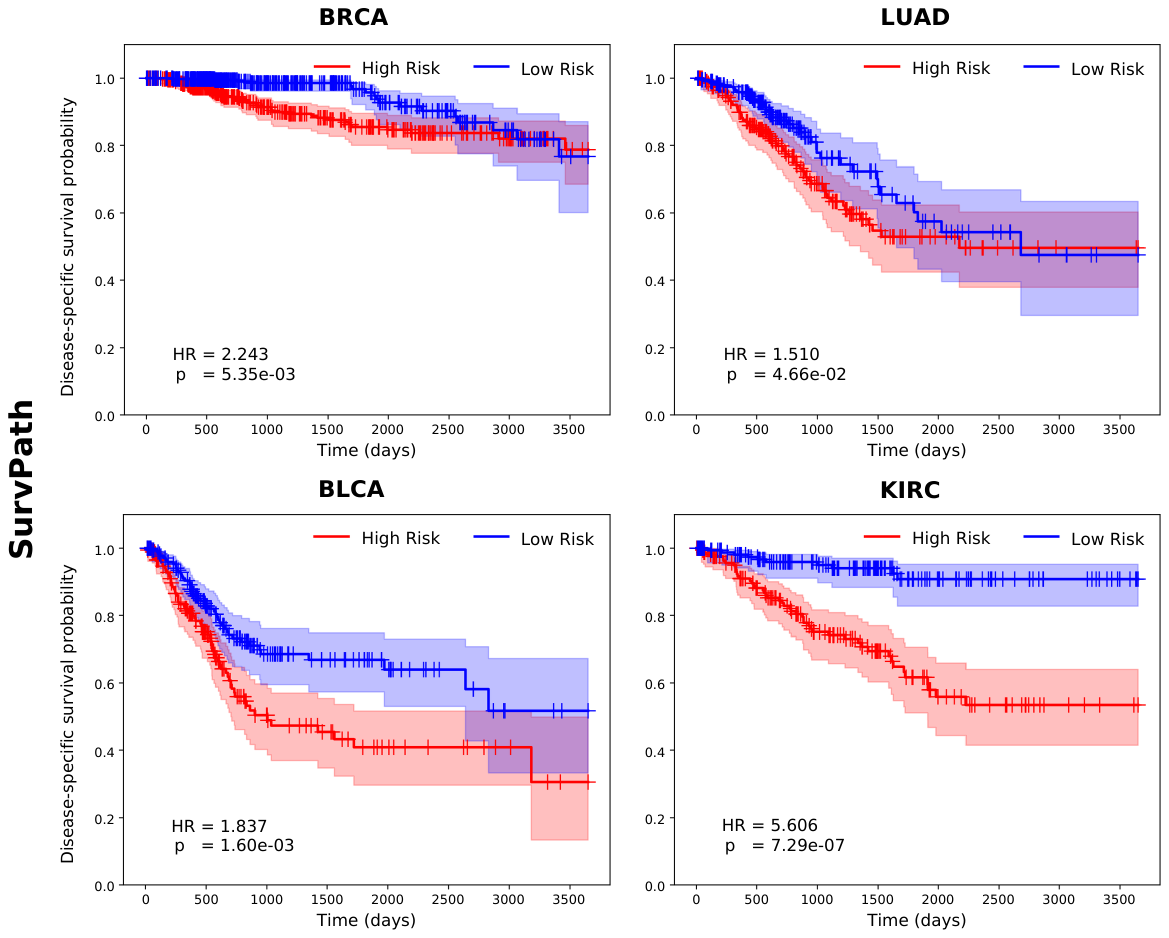}
    \caption{Kaplan–Meier survival curves for SurvPath, showing high- and low-risk patient groups stratified by the median predicted risk score, with corresponding hazard ratios and p-values.}
    \label{fig:km_analysis_survpath}
\end{figure}

\begin{figure}[H]
    \centering
        \includegraphics[width=\linewidth, trim=0 380 0 0]{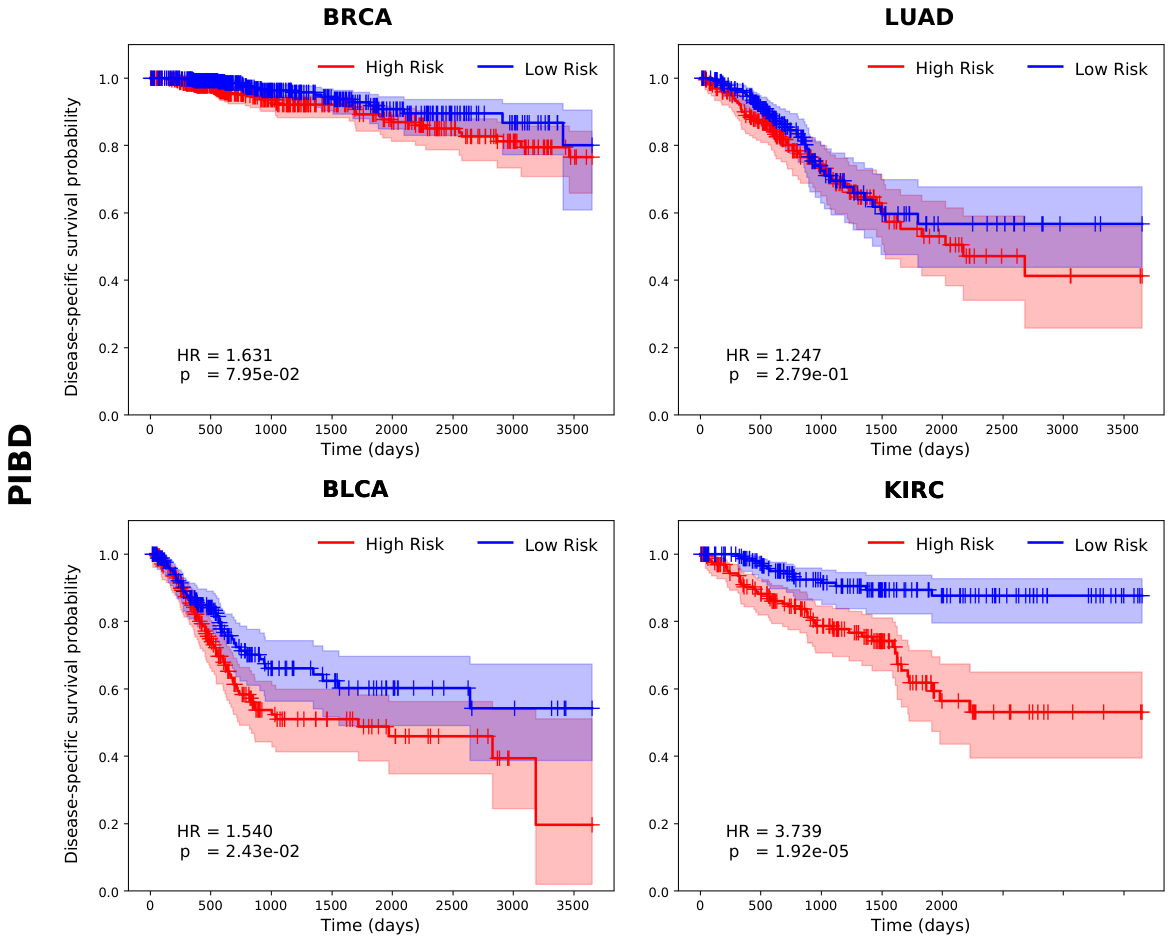}
    \caption{Kaplan–Meier survival curves for PIBD, showing high- and low-risk patient groups stratified by the median predicted risk score, with corresponding hazard ratios and p-values.}
    \label{fig:km_analysis_PIBD}
\end{figure}

\begin{table}[H]
\caption{Disentanglement test results measured by the orthogonal score over the different folds (mean $\pm$ std). We denote the best and second-best total disentanglement by \textbf{bold} and \underline{underlined}, respectively.}\label{tab2}%
\begin{tabular}{lllllll}
\toprule
& & \textbf{BRCA}  & \textbf{BLCA} & \textbf{LUAD} & \textbf{KIRC} & \textbf{Avg. }\\
\midrule
PIBD & D1    & 0.334±0.024 & 0.216±0.055 & 0.246±0.116 & 0.183±0.063 & 0.245 \\
                & D2    & 0.159±0.052 & 0.367±0.086 & 0.354±0.072 & 0.328±0.021 & 0.302 \\
                & Total & 0.246±0.034 & 0.291±0.051 & 0.300±0.068 & 0.256±0.022 & 0.273 \\
  \midrule
DIMAF & D1    & 0.073±0.027 & 0.083±0.020 & 0.090±0.038 & 0.069±0.034 & 0.079 \\
                  & D2    & 0.060±0.018 & 0.051±0.009 & 0.042±0.007 & 0.042±0.006 & 0.049 \\
                  & Total & \underline{0.066±0.009} & 0.067±0.012 & 0.066±0.020 & \underline{0.056±0.015} & \underline{0.064} \\
    \midrule
DIMAFx$_{\text{nodis}}$ & D1    & 0.070±0.019 & 0.067±0.011 & 0.062±0.007 & 0.063±0.011 & 0.066 \\
                        & D2    & 0.049±0.007 & 0.050±0.017 & 0.047±0.010 & 0.051±0.018 & 0.049 \\
                        & Total & \textbf{0.060±0.007} & \textbf{0.058±0.006} & \textbf{0.055±0.003} & 0.057±0.011 & \textbf{0.058} \\
  \midrule
DIMAFx  & D1    & 0.071±0.037 & 0.087±0.038 & 0.073±0.024 & 0.053±0.017 & 0.071 \\
        & D2    & 0.050±0.011 & 0.045±0.013 & 0.042±0.008 & 0.044±0.006 & 0.045 \\
        & Total & \textbf{0.060±0.016} & \underline{0.066±0.020} & \underline{0.057±0.012} & \textbf{0.048±0.006} & \textbf{0.058} \\
\bottomrule
\end{tabular}
\label{A:OS_res}
\end{table}

\begin{table}[H]
\caption{Normalized SHAP values of the disentangled representations over the different folds (mean $\pm$ std).}\label{tab2}%
\begin{tabular}{lllllll}
\toprule
 & & \textbf{BRCA}  & \textbf{BLCA} & \textbf{LUAD} & \textbf{KIRC} & \textbf{Avg. }\\
\midrule
  DIMAF & Specific & 0.185±0.038 & 0.273±0.038 & 0.257±0.027 & 0.302±0.043 & 0.254 \\
 & Shared & 0.815±0.038 & 0.727±0.038 & 0.743±0.027 & 0.698±0.043 & 0.746 \\
  \midrule
DIMAFx$_{\text{nodis}}$ & Specific & 0.543±0.028 & 0.569±0.022 & 0.561±0.027 & 0.603±0.034 & 0.569 \\
 & Shared & 0.457±0.028 & 0.431±0.022 & 0.439±0.027 & 0.397±0.034 & 0.431 \\
  \midrule
  DIMAFx & Specific & 0.168±0.025 & 0.267±0.023 & 0.235±0.017 & 0.313±0.054 & 0.246 \\
 & Shared & 0.832±0.025 & 0.733±0.023 & 0.765±0.017 & 0.687±0.054 & 0.754 \\
\bottomrule
\end{tabular}
\label{A:shap_general}
\end{table}

\begin{figure}[H]
    \centering
\includegraphics[width=\textwidth, trim=0 510 0 0]{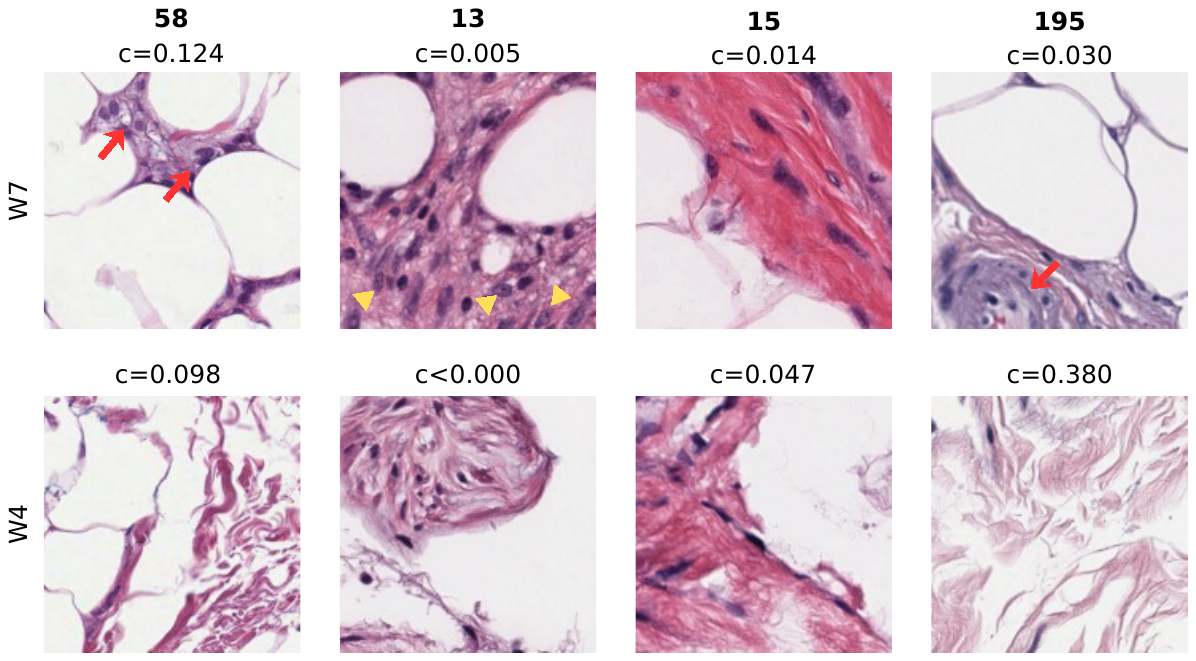}
    \caption{Visualization of the ``W7: Adipose $>$ stroma" and ``W4: Adipose $>$ stroma" prototype features for four representative cases, showing fibrous and fibroadipose tissue. \textbf{Case 58}: the W7 patch shows adipose tissue with histiocytic infiltration (arrows), suggestive of fat necrosis, while the W4 patch shows fibroadipose tissue without apparent abnormality. \textbf{Case 13}: the W7 patch demonstrates fibroadipose tissue with activated fibroblasts (arrowheads) and a mild chronic inflammatory infiltrate; the W4 patch shows fibrous tissue fragments without apparent abnormality. \textbf{Case 15}: the W7 patch shows fibroadipose tissue with unremarkable findings, and the W4 patch shows fibrous tissue. \textbf{Case 195}: the W7 patch demonstrates fibroadipose tissue with a blood vessel apparent within the fibrous tissue (arrow), whereas the W4 patch shows detached collagen fibers within the stroma.}
\label{fig:multimodalinterpretation_W7} 
\end{figure}

\begin{figure}[H]
    \centering
    \includegraphics[width=\textwidth, trim=0 300 0 0]{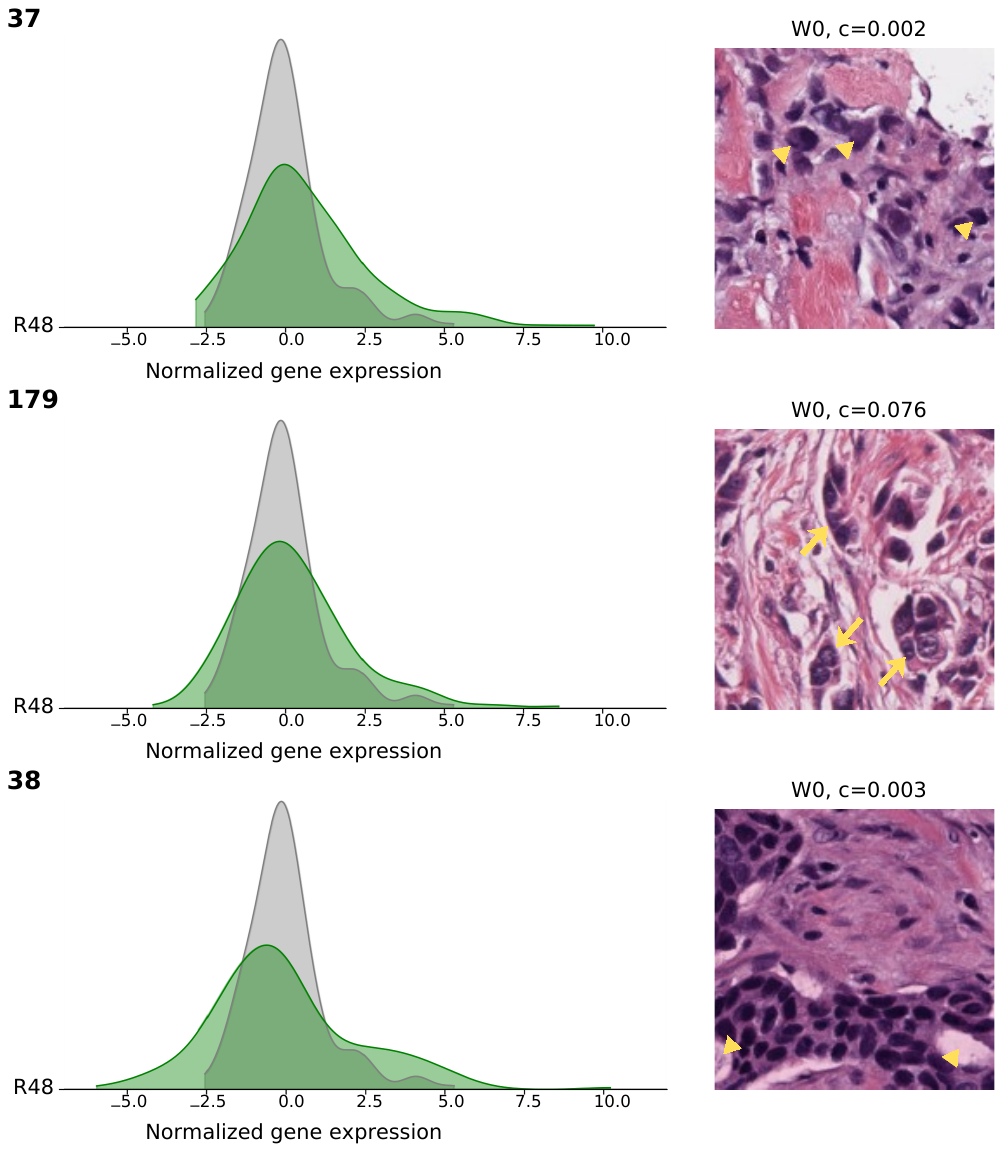}
    \caption{Visualization of the ``R48: KRAS signaling down" and ``W0: Tumor" features for three representative cases. The ridge plots show the frequency distributions of log transformed RSEM normalized gene expression values for the ``R48: KRAS signaling down" pathway in the highlighted case (green) and averaged over all train samples (gray). For the ``W0: Tumor" feature, we show the prototype cardinality (c) and a representative patch for each case. \textbf{Case 37}: the W0 prototype has a relatively low cardinality and shows tumor cells infiltrating singly within stroma and exhibiting hyperchromatic (darkly stained), atypical nuclei (arrowheads). This is consistent with a high-grade morphology. The KRAS signaling down pathway distribution seems slightly upregulated. \textbf{Case 179}: W0 shows atypical tumor cell infiltration forming small nests (arrows) with high-grade nuclear atypia. This is a high-grade tumor with a high frequency. Despite this, the corresponding SHAP value indicates a decrease in risk to the model’s prediction, and the case was assigned a low-risk score. Here, the pathway distribution is very similar to the baseline distribution. \textbf{Case 38}: W0 shows tumor cells forming tubular structures (arrowheads) with intermediate nuclear atypia and hyperchromasia. This feature exhibits an intermediate-grade morphology. The R48 feature shows a slightly downregulated pathway distribution, which was assigned a relatively high SHAP value.}
\label{fig:multimodalinterpretation_R48} 
\end{figure}




\begin{table}[H]
\centering
\caption{Transcriptomic pathway features annotations from the Molecular Signatures Database (MSigDB) hallmark gene set collection, part 1.}
\label{A:RNAfeatspt1}

{\small
\begin{tabularx}{\textwidth}{l|>{\raggedright\arraybackslash}X}
\textbf{Feature name} & \textbf{Short description} \\
\hline
R0: TNFA signaling via NFKB & Genes regulated by NF-kB in response to TNF \\
R1: Hypoxia & Genes up-regulated in response to low oxygen levels (hypoxia)\\
R2: Cholesterol homeostasis & Genes involved in cholesterol homeostasis \\
R3: Mitotic spindle & Genes important for mitotic spindle assembly \\
R4: WNT beta catenin signaling & Genes up-regulated by activation of WNT signaling through accumulation of beta catenin CTNNB1 \\
R5: TGF beta signaling & Genes up-regulated in response to TGFB1 \\
R6: IL6 JAK STAT3 signaling & Genes up-regulated by IL6 via STAT3, \textit{e.g.}, during acute phase response \\
R7: DNA repair & Genes involved in DNA repair \\
R8: G2M checkpoint & Genes involved in the G2/M checkpoint, as in progression through the cell division cycle \\
R9: Apoptosis & Genes mediating programmed cell death (apoptosis) by activation of caspases \\
R10: Notch signaling & Genes up-regulated by activation of Notch signaling \\
R11: Adipogenesis & Genes up-regulated during adipocyte differentiation (adipogenesis) \\
R12: Estrogen response early & Genes defining early response to estrogen \\
R13: Estrogen response late & Genes defining late response to estrogen \\
R14: Androgen response & Genes defining response to androgens \\
R15: Myogenesis & Genes involved in development of skeletal muscle (myogenesis) \\
R16: Protein secretion & Genes involved in protein secretion pathways \\
R17: Interferon alpha response & Genes up-regulated in response to alpha interferon proteins \\
R18: Interferon gamma response & Genes up-regulated in response to IFNG \\
R19: Apical junction & Genes encoding components of apical junction complex \\
R20: Apical surface & Genes encoding proteins over-represented on the apical surface of epithelial cells, e.g., important for cell polarity (apical area) \\
R21: Hedgehog signaling & Genes up-regulated by activation of hedgehog signaling \\
R22: Complement & Genes encoding components of the complement system, which is part of the innate immune system \\
R23: Unfolded protein response & Genes up-regulated during unfolded protein response, a cellular stress response related to the endoplasmic reticulum \\
R24: PI3K AKT MTOR signaling & Genes up-regulated by activation of the PI3K/AKT/mTOR pathway \\
\end{tabularx}
}
\end{table}

\begin{table}[h!]
\centering
\caption{Transcriptomic pathway features annotations from the Molecular Signatures Database (MSigDB) hallmark gene set collection, part 2.}
\label{A:RNAfeatspt2}

{\small
\begin{tabularx}{\textwidth}{l|>{\raggedright\arraybackslash}X}
\textbf{Feature name} & \textbf{Short description} \\
\hline
R25: MTORC1 signaling & Genes up-regulated through activation of mTORC1 complex \\
R26: E2F targets & Genes encoding cell cycle related targets of E2F transcription factors \\
R27: MYC targets v1 & A subgroup of genes regulated by MYC - version 1 (v1) \\
R28: MYC targets v2 & A subgroup of genes regulated by MYC - version 2 (v2) \\
R29: Epithelial mesenchymal transition & Genes defining epithelial-mesenchymal transition, as in wound healing, fibrosis and metastasis \\
R30: Inflammatory response & Genes defining inflammatory response \\
R31: Xenobiotic metabolism & Genes encoding proteins involved in processing of drugs and other xenobiotics \\
R32: Fatty acid metabolism & Genes encoding proteins involved in metabolism of fatty acids \\
R33: Oxidative phosphorylation & Genes encoding proteins involved in oxidative phosphorylation \\
R34: Glycolysis & Genes encoding proteins involved in glycolysis and gluconeogenesis \\
R35: Reactive oxygen species pathway & Genes up-regulated by reactive oxigen species (ROS) \\
R36: P53 pathway & Genes involved in p53 pathways and networks \\
R37: UV response up & Genes up-regulated in response to ultraviolet (UV) radiation \\
R38: UV response down & Genes down-regulated in response to ultraviolet (UV) radiation \\
R39: Angiogenesis & Genes up-regulated during formation of blood vessels (angiogenesis) \\
R40: Heme metabolism & Genes involved in metabolism of heme (a cofactor consisting of iron and porphyrin) and erythroblast differentiation \\
R41: Coagulation & Genes encoding components of blood coagulation system; also up-regulated in platelets \\
R42: IL2 STAT5 signaling & Genes up-regulated by STAT5 in response to IL2 stimulation\\
R43: Bile acid metabolism & Genes involve in metabolism of bile acids and salts \\
R44: Peroxisome & Genes encoding components of peroxisome \\
R45: Allograft rejection & Genes up-regulated during transplant rejection \\
R46: Spermatogenesis & Genes up-regulated during production of male gametes (sperm), as in spermatogenesis \\
R47: KRAS signaling up & Genes up-regulated by KRAS activation \\
R48: KRAS signaling down & Genes down-regulated by KRAS activation \\
R49: Pancreas beta cells & Genes specifically up-regulated in pancreatic beta cells \\
\end{tabularx}
}
\end{table}

\begin{table}[h!]
\caption{WSI feature annotations for BRCA fold 2, showing the prototype annotation and the cardinality (Card.) of each prototype feature for the train and test set. For a given annotation with multiple morphologies ($X$ and $Y$), a notation of $X >> Y$ indicates that $X$ is strongly dominant relative to $Y$, whereas $X > Y$ denotes that $X$ remains dominant, but $Y$ is more prevalent compared to with $X >> Y$.}
\label{A:wsifeats}

{\small
\begin{tabularx}{\textwidth}{l|>{\raggedright\arraybackslash}X|l|>{\raggedright\arraybackslash}X|l}
 & \multicolumn{2}{l|}{\textbf{Train set}} & \multicolumn{2}{l}{\textbf{Test set}} \\
\hline
\textbf{Feature} & \textbf{Annotation} & \textbf{Card.} & \textbf{Annotation} & \textbf{Card.} \\
\hline
W0  & Tumor & 0.0951 & Tumor & 0.062689 \\
W1  & Tissue edge, stroma & 0.0501 & -- & 0.000018 \\
W2  & Tissue edge, stroma $>$ tumor & 0.0212 & Tissue edge, stroma $>$ tumor & 0.000033 \\
W3  & Tumor & 0.0773 & Tumor & 0.009016\\
W4  & Adipose $>$ stroma & 0.0620 & Adipose $>$ stroma & 0.098484\\
W5  & Tumor & 0.0402 & -- & 0.000020\\
W6  & Stroma $>$ adipose & 0.0543 & Stroma $>$ adipose & 0.065152\\
W7  & Adipose $>$ stroma & 0.0529 & Adipose $>$ stroma & 0.098530 \\
W8  & Tumor, solid pattern, abundant cytoplasm & 0.0725 & Tumor, solid pattern, abundant cytoplasm & 0.197805 \\
W9  & Tumor, low grade $>>$ adipose & 0.0715 & Tumor $>>$ adipose & 0.107616\\
W10 & Stroma $>$ tumor $>$ immune cells & 0.0391 & Stroma $>$ tumor $>$ immune cells & 0.053608\\
W11 & Tissue edge, mixed $>$ immune cells, stroma & 0.0840 & Tumor $>$ stroma $>$ immune cells & 0.012783 \\
W12 & Normal ducts/lobules & 0.0301 & Normal ducts/lobules & 0.035938\\
W13 & Stroma $>$ immune cells & 0.0636 & Stroma $>$ adipose & 0.005866\\
W14 & Edge stainings & 0.0967 & Edge stainings, blood vessels & 0.118145 \\
W15 & Stroma $>$ immune cells & 0.0897 & Stroma $>$ immune cells & 0.134299 \\
\end{tabularx}
}
\end{table}

\end{document}